%% file: main.tex
\documentclass[acmtog,authorversion,nonacm]{acmart}
\AtBeginDocument{%
  \providecommand\BibTeX{{%
    \normalfont B\kern-0.5em{\scshape i\kern-0.25em b}\kern-0.8em\TeX}}}

\setcopyright{none}
\copyrightyear{2024}
\acmYear{2024}
\acmDOI{XXXXXXX.XXXXXXX}

\usepackage{savesym}
\savesymbol{zifour@default}
\savesymbol{zifour@scaled}
\usepackage{enumerate}
\usepackage{multirow} 
\usepackage{array}
\usepackage{cleveref}
\usepackage{inconsolata}
\usepackage{spverbatim}
\usepackage{wrapfig}

\usepackage{graphicx}

\input{macros.tex}

\begin{document}

\title[Open-Universe Indoor Scene Generation]{Open-Universe Indoor Scene Generation using\\LLM Program Synthesis and Uncurated Object Databases}

\author{Rio Aguina-Kang}
\authornote{These authors contributed equally}
\email{raguinakang@ucsd.edu}
\affiliation{%
    \institution{UC San Diego}
    \country{USA}
}
\author{Maxim Gumin}
\authornotemark[1]
\email{maxgumin@gmail.com}
\affiliation{%
    \institution{Brown University}
    \country{USA}
}
\author{Do Heon Han}
\authornotemark[1]
\email{do\_heon\_han@brown.edu}
\affiliation{%
    \institution{Brown University}
    \country{USA}
}
\author{Stewart Morris}
\authornotemark[1]
\email{stewart\_morris@brown.edu}
\affiliation{%
    \institution{Brown University}
    \country{USA}
}
\author{Seung Jean Yoo}
\authornotemark[1]
\email{seung\_jean\_yoo@brown.edu}
\affiliation{%
    \institution{Brown University}
    \country{USA}
}
\author{Aditya Ganeshan}
\email{aditya\_ganeshan@brown.edu}
\affiliation{%
    \institution{Brown University}
    \country{USA}
}
\author{R. Kenny Jones}
\email{russell_jones@brown.edu}
\affiliation{%
    \institution{Brown University}
    \country{USA}
}
\author{Qiuhong Anna Wei}
\email{qiuhong\_wei@brown.edu}
\affiliation{%
    \institution{Brown University}
    \country{USA}
}
\author{Kailiang Fu}
\email{kailiang.fu@dymaxion.design}
\affiliation{%
    \institution{Dymaxion, LLC}
    \country{USA}
}
\author{Daniel Ritchie}
\email{daniel\_ritchie@brown.edu}
\affiliation{%
    \institution{Brown University}
    \country{USA}
}

\begin{abstract}
\input{00-abstract.tex}

\end{abstract}

\ccsdesc[500]{Computing methodologies~Computer graphics}
\ccsdesc[500]{Computing methodologies~Neural networks}
\ccsdesc[500]{Computing methodologies~Natural language generation}

\keywords{indoor scene synthesis, program synthesis, layout generation, large language models, vision language models, foundation models}

\begin{teaserfigure}
    \centering
    \setlength{\tabcolsep}{2pt}
    \newcolumntype{y}{>{\centering\arraybackslash}p{0.24\linewidth}}
    \begin{tabular}{yyyy}
         \textit{``A living room for watching TV''} &
         \textit{``A high-end mini restaurant''} &
         \textit{``A witch's room with a cauldron'} &
         \textit{``A Japanese living room''}
         \\
         \includegraphics[width=\linewidth]{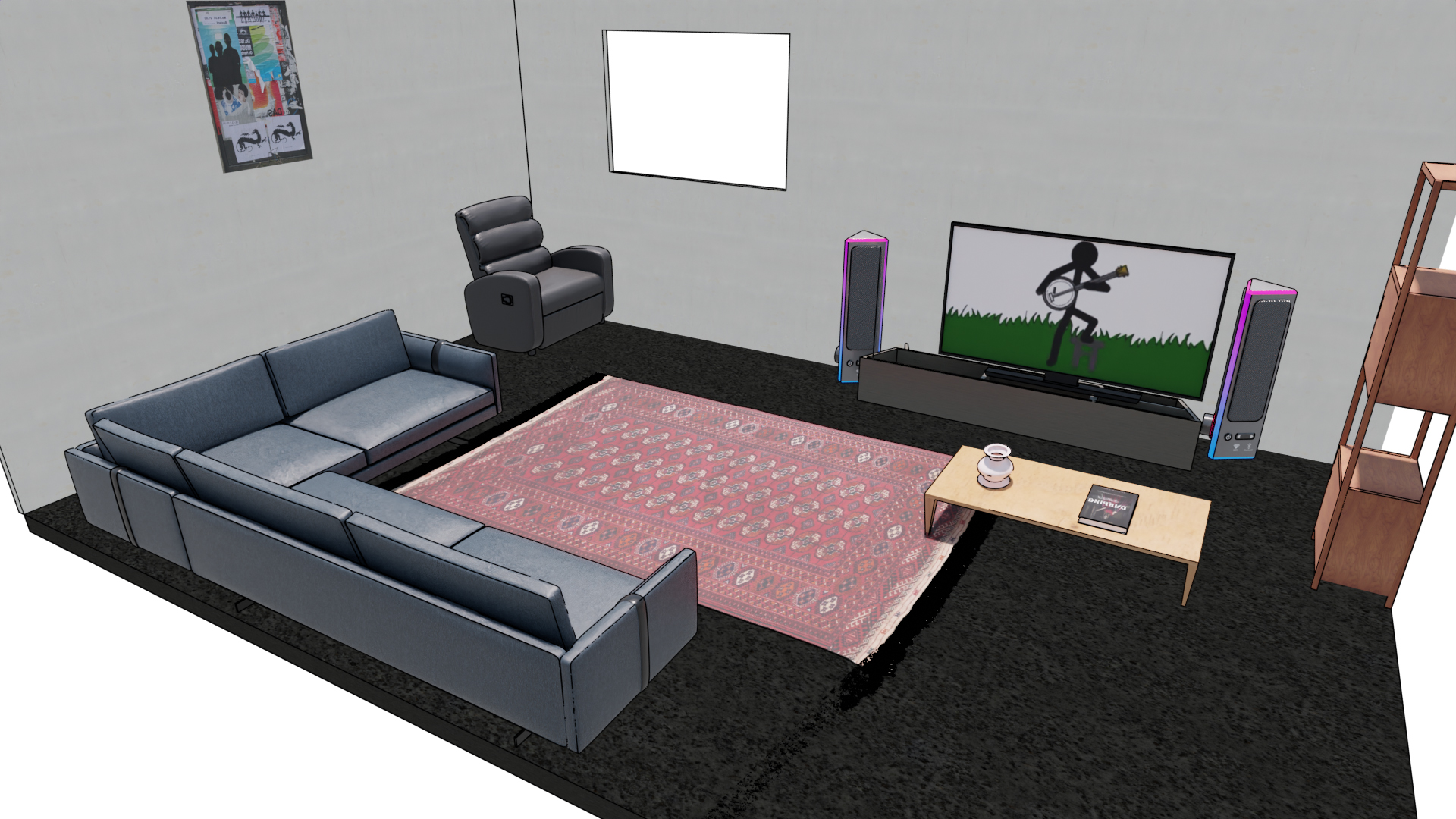} &
         \includegraphics[width=\linewidth]{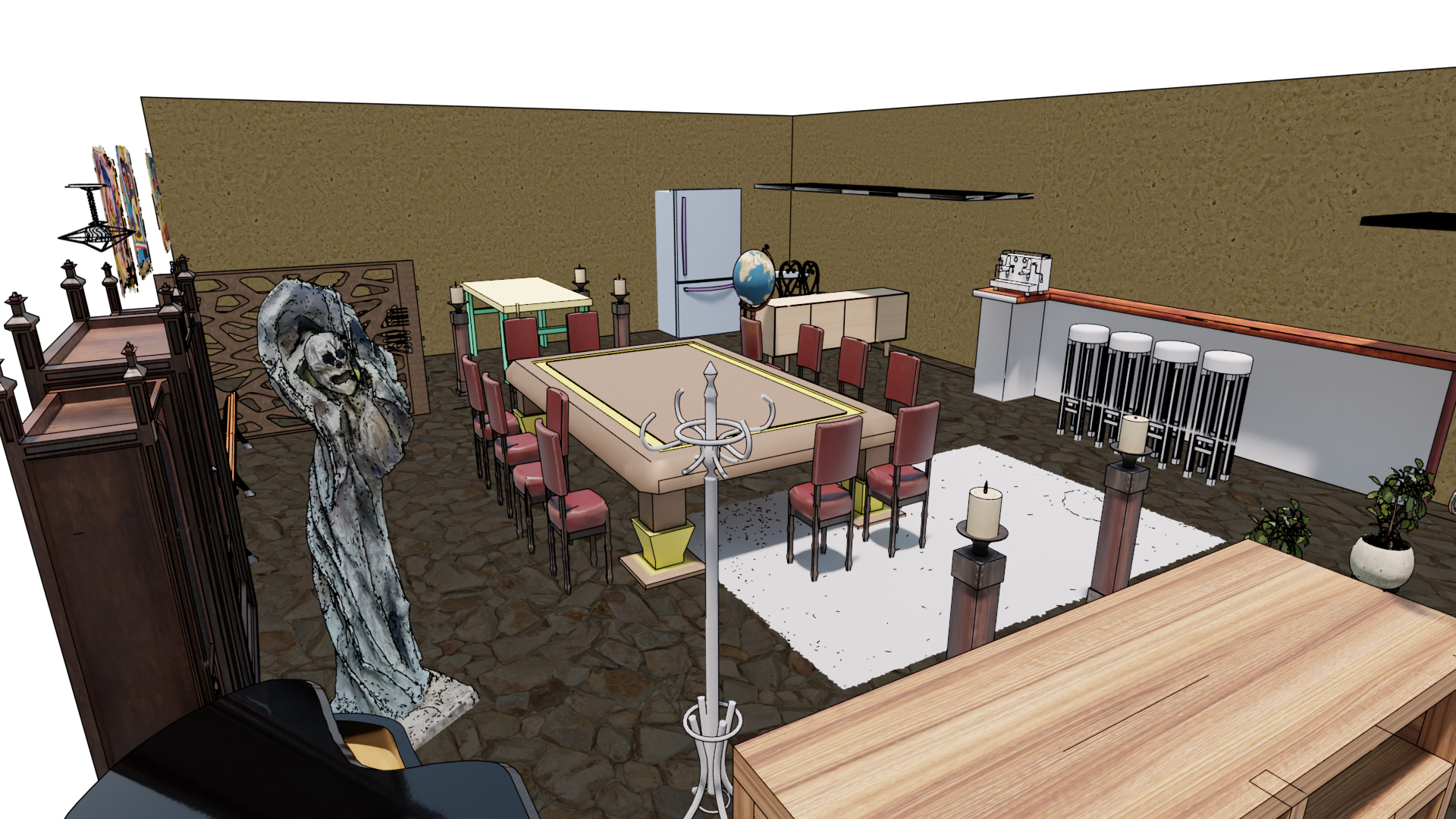} &
         \includegraphics[width=\linewidth]{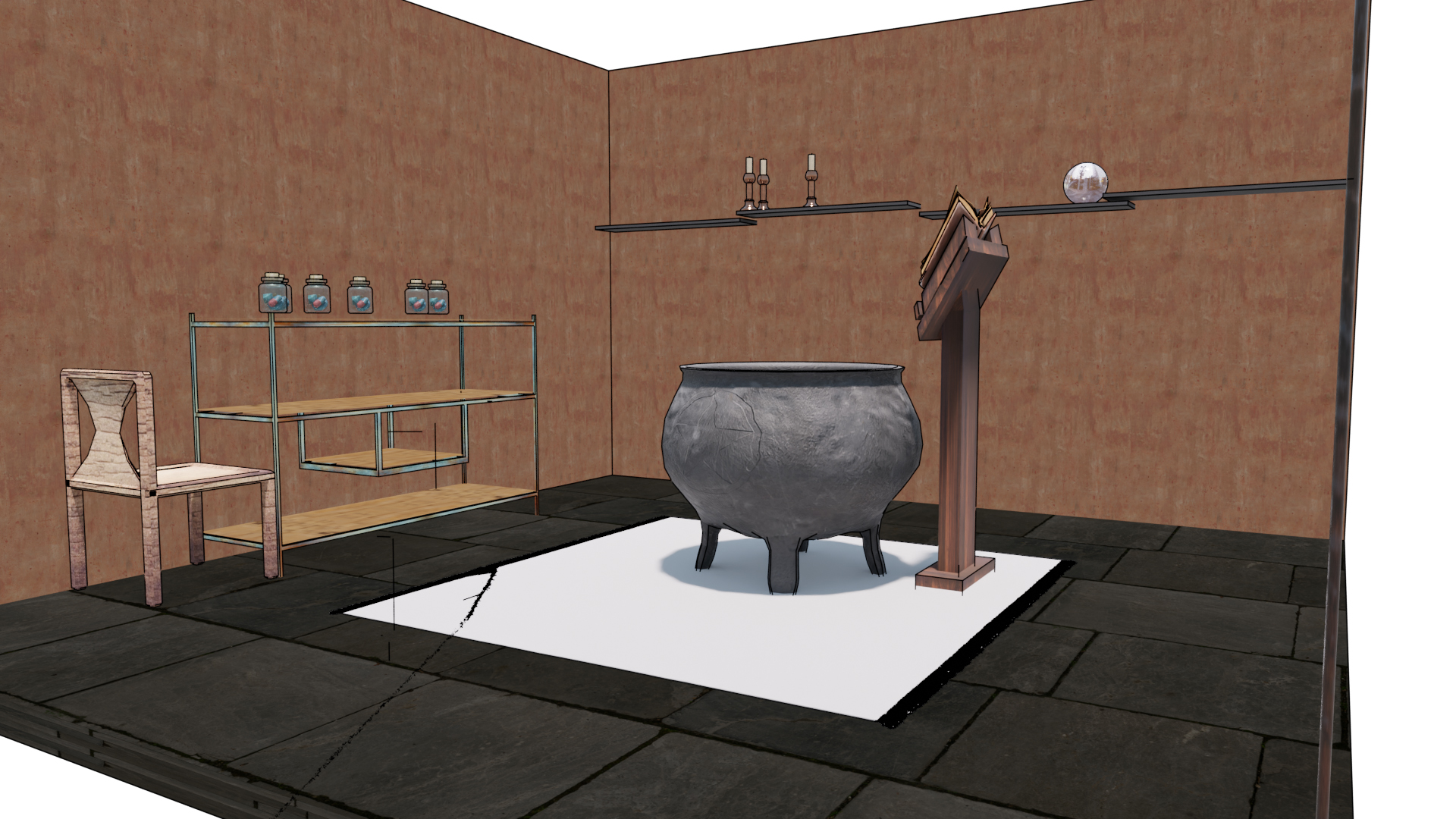} &
         \includegraphics[width=\linewidth]{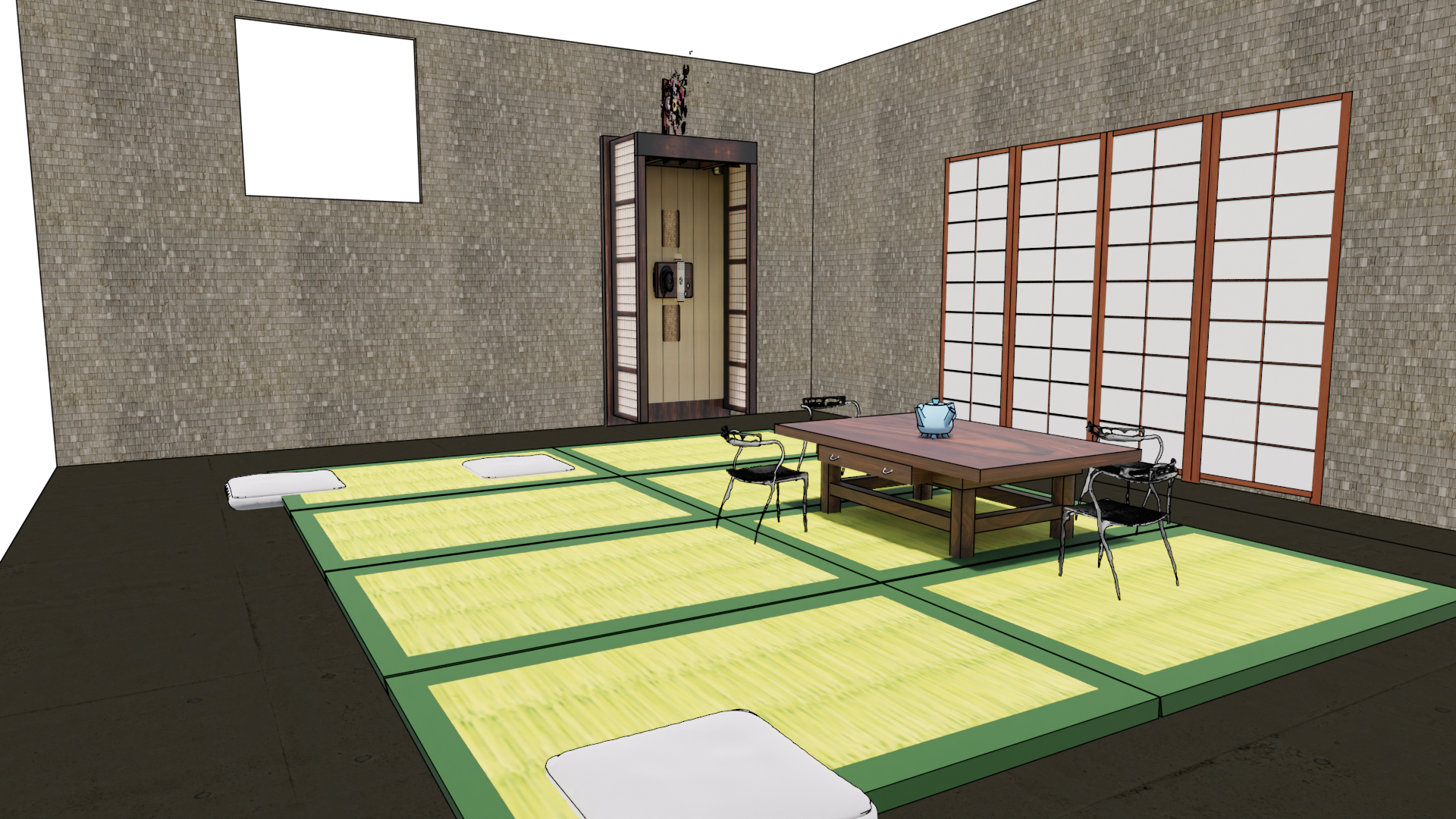}
         \\
         \textit{``A living room''} &
         \textit{``A dining room for one''} &
         \textit{``A bedroom''} &
         \textit{``An old-fashioned bedroom''}
         \\
         \includegraphics[width=\linewidth]{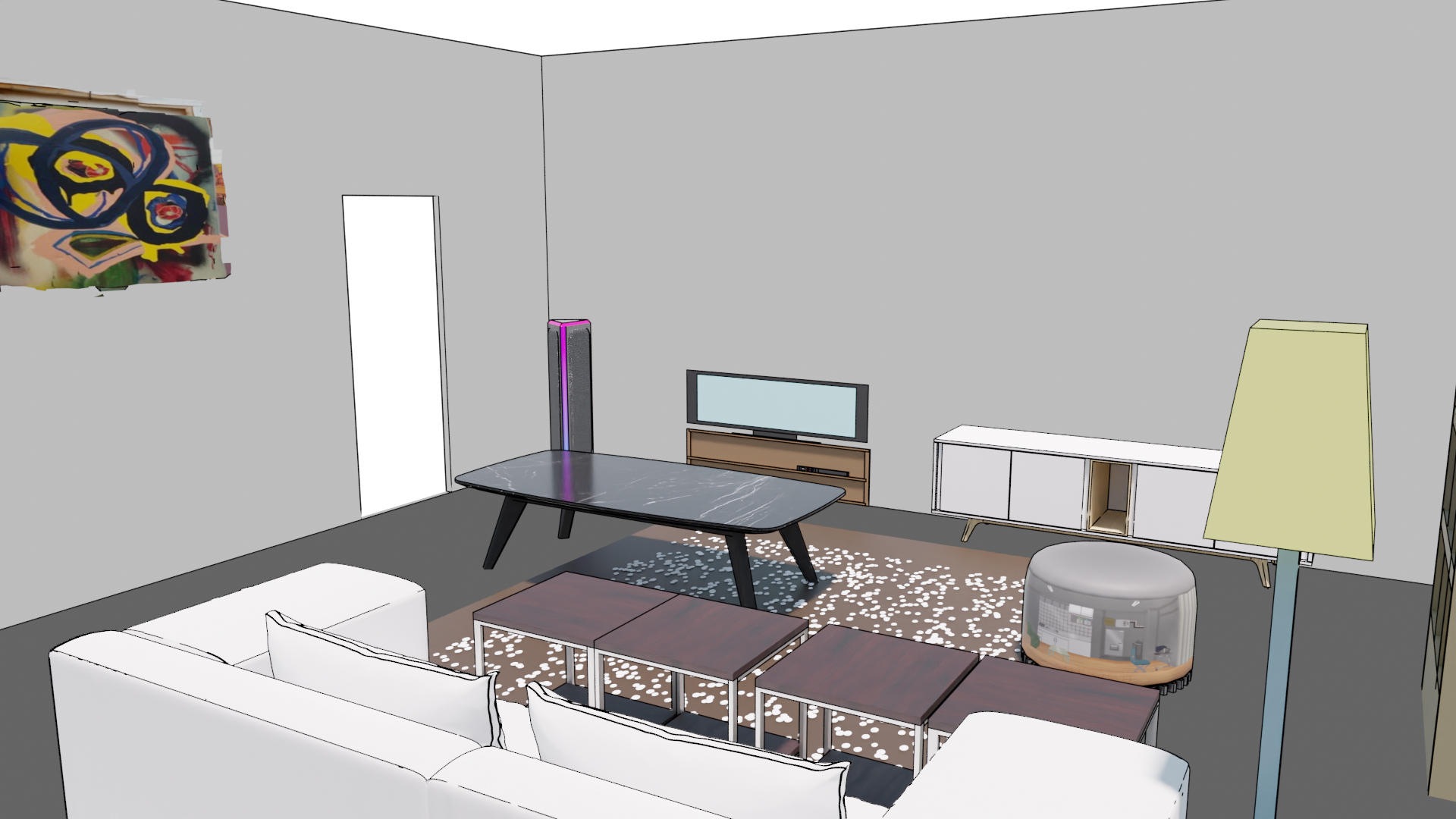} &
         \includegraphics[width=\linewidth]{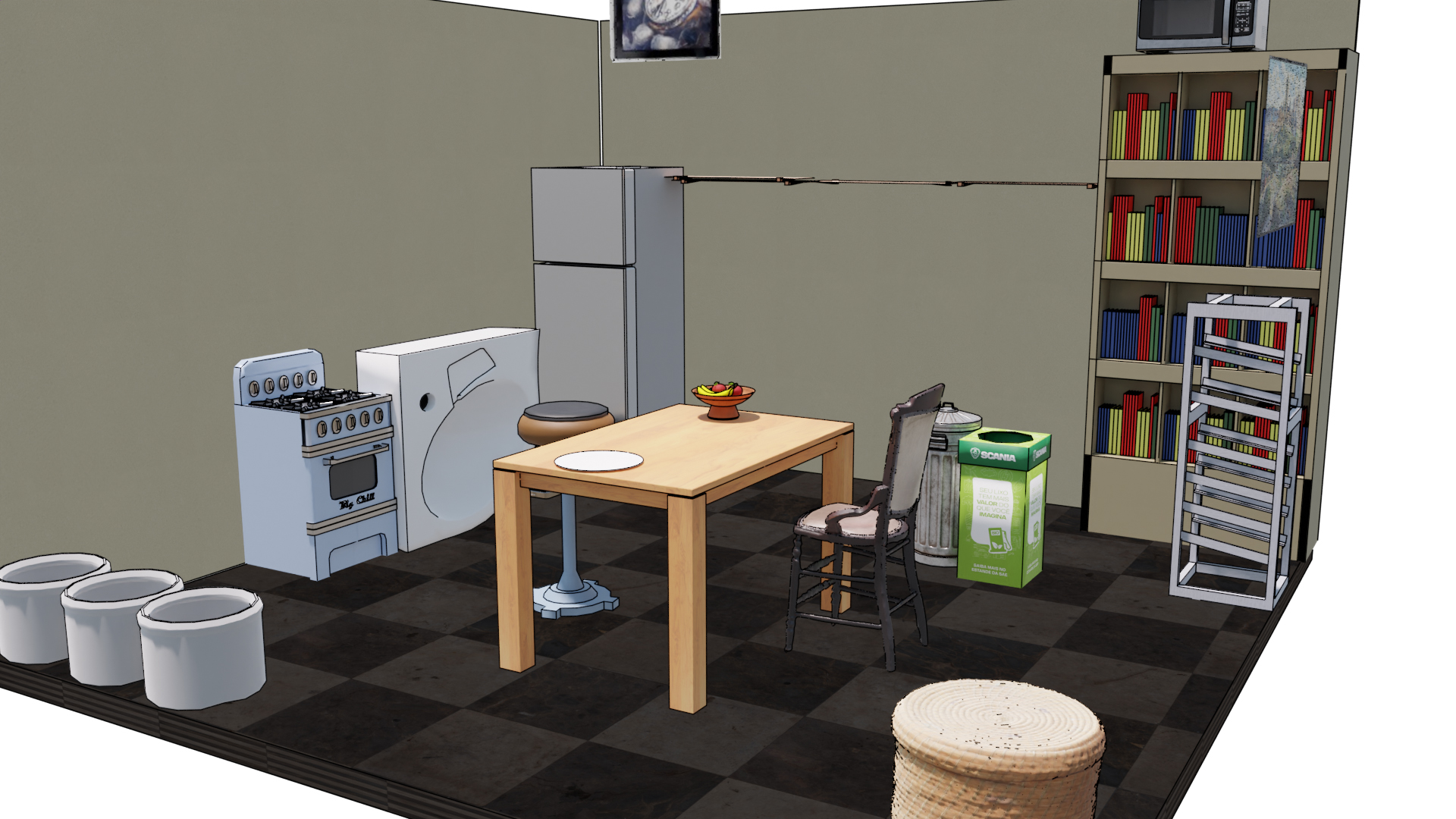} &
         \includegraphics[width=\linewidth]{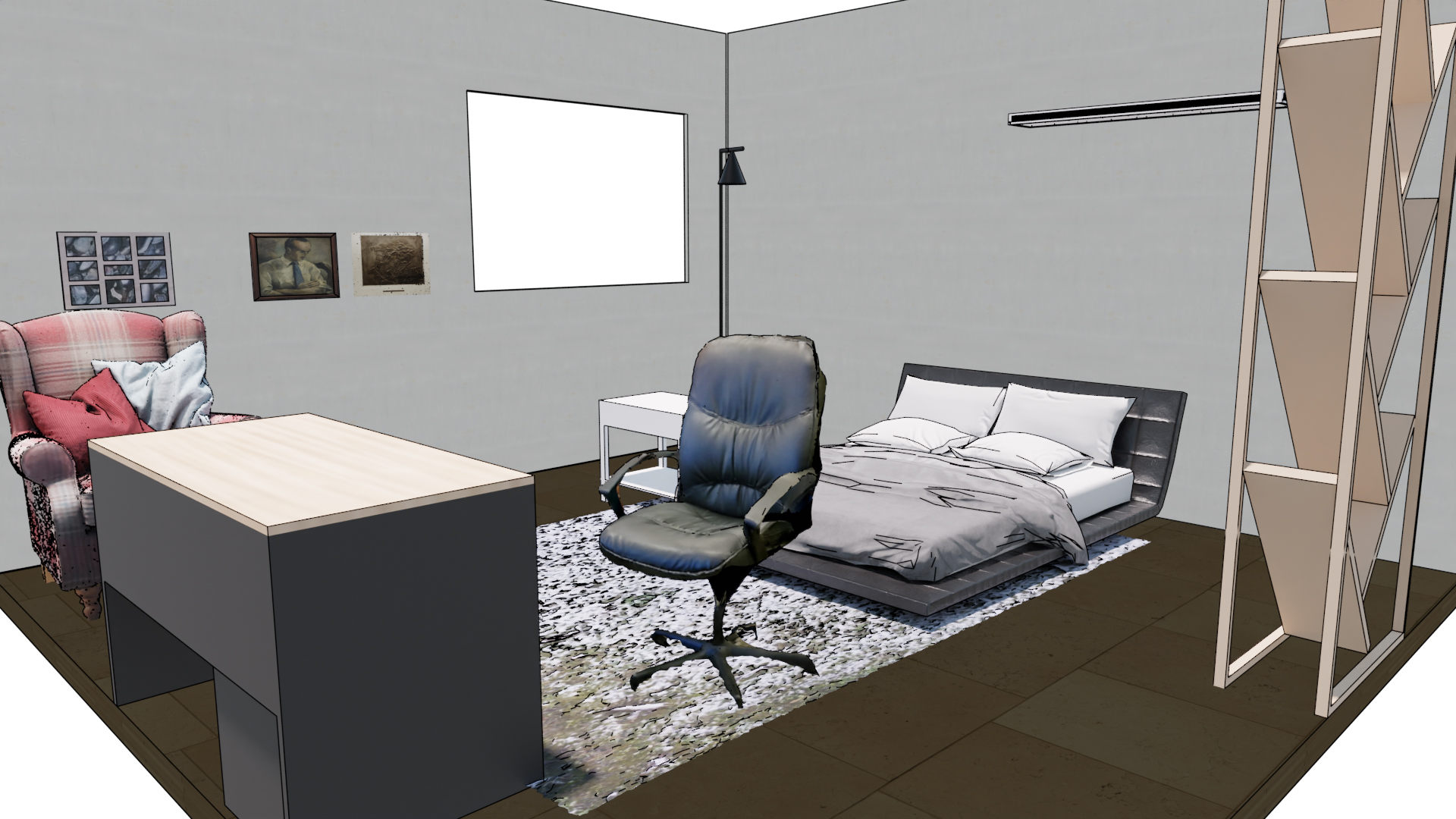} &
         \includegraphics[width=\linewidth]{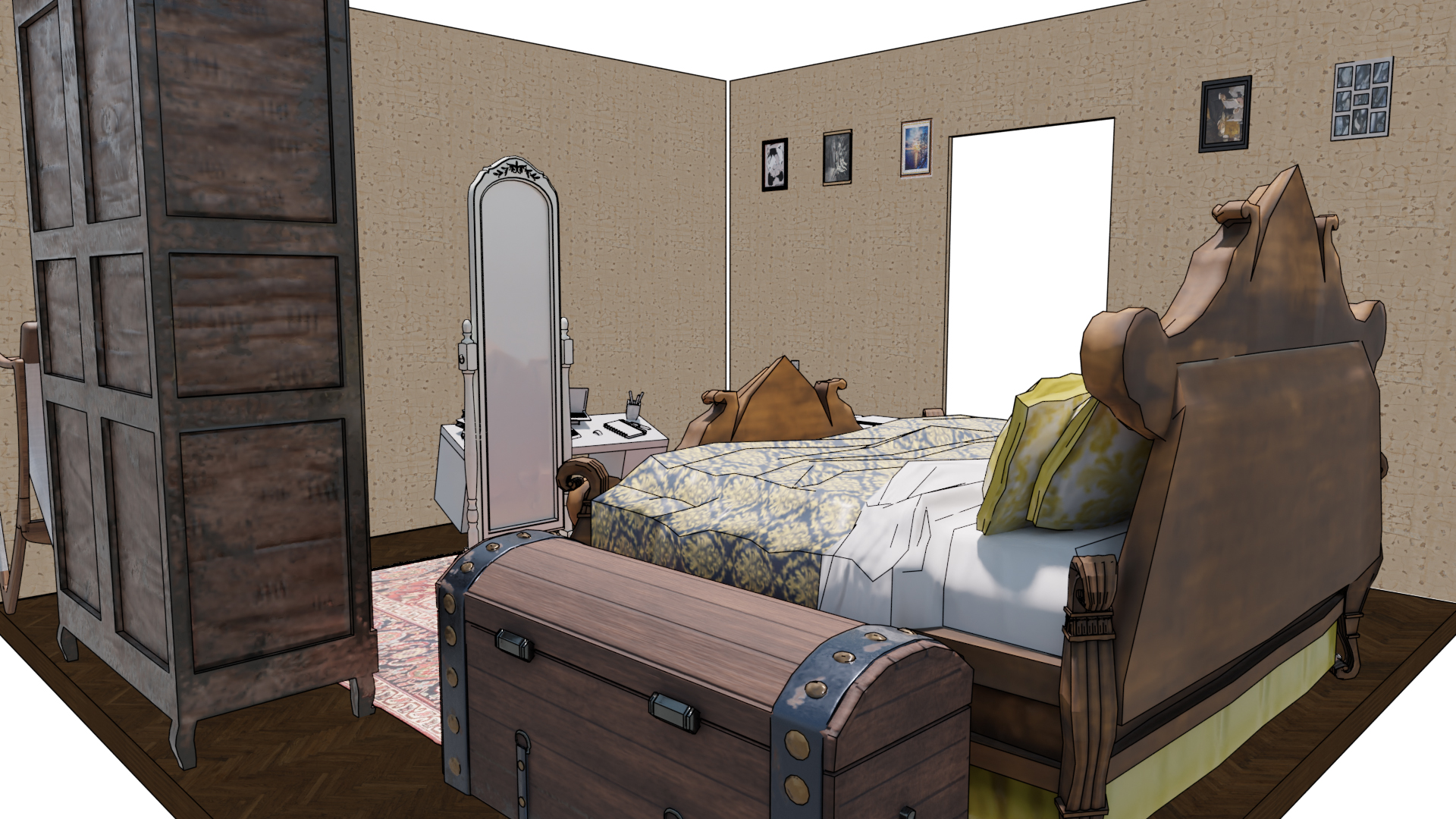}
    \end{tabular}
    \caption{
        Our method generates 3D indoor scenes from open-ended text prompts.
        Generated scenes are not limited to a fixed set of room types or object categories; this ``open-universe'' capability is enabled by judicious use of pre-trained large language models (LLMs) and vision-language models (VLMs).
    }
    \label{fig:teaser}
\end{teaserfigure}

\maketitle

\input{01-intro.tex}
\input{02-related.tex}

\input{03-overview.tex}
\input{04-language.tex}

\input{05-progsynth.tex}
\input{06-layoutopt.tex}

\input{07-retrievealign.tex}

\input{08-results.tex}
\input{09-conclusion.tex}

\bibliographystyle{ACM-Reference-Format}
\bibliography{main}
\input{10-appendix.tex}

\end{document}

%% file: macros.tex
\usepackage{xspace}

\newcommand{\relu}{\mathrm{relu}}

\newcommand{\proj}{\mathrm{proj}}

\newcommand{\revdel}[1]{}

\newcommand{\denselist}{\itemsep 0pt\parsep=0pt\partopsep 0pt\vspace{-\topsep}}



%% file: 00-abstract.tex
We present a system for generating indoor scenes in response to text prompts.
The prompts are not limited to a fixed vocabulary of scene descriptions, and the objects in generated scenes are not restricted to a fixed set of object categories---we call this setting \emph{open-universe} indoor scene generation.
Unlike most prior work on indoor scene generation, our system does not require a large training dataset of existing 3D scenes.
Instead, it leverages the world knowledge encoded in pre-trained large language models (LLMs) to synthesize programs in a domain-specific layout language that describe objects and spatial relations between them.
Executing such a program produces a specification of a constraint satisfaction problem, which the system solves using a gradient-based optimization scheme to produce object positions and orientations.
To produce object geometry, the system retrieves 3D meshes from a database.
Unlike prior work which uses databases of category-annotated, mutually-aligned meshes, we develop a pipeline using vision-language models (VLMs) to retrieve meshes from massive databases of un-annotated, inconsistently-aligned meshes.
Experimental evaluations show that our system outperforms generative models trained on 3D data for traditional, closed- universe scene generation tasks; it also outperforms a recent LLM-based layout generation method on open-universe scene generation.

%% file: 01-intro.tex
\section{Introduction}

Many people spend a significant portion of their lives indoors: in their homes, workplaces, social gathering spaces, etc. 
Unsurprisingly, indoor environments also feature heavily in virtual depictions of the real world: in games, extended reality experiences, and architectural visualizations.
Such virtual scenes have real-world uses, as well.
For example, there are now a variety of free-to-use interior design tools online which allow users to explore virtual re-designs of their own real spaces~\cite{Planner5d,RoomSketcher,targetHomePlanner}.
In addition, furniture and home product retailers are increasingly using renderings of virtual scenes to stage and advertise their products, as the process of doing so is easier, less expensive, and more adaptable to different regions of the world than taking physical photographs~\cite{IKEARendering}.
Finally, virtual indoor scenes have become a critical data source for training autonomous embodied agents to perceive and navigate within typical indoor environments~\cite{proc_thor, puig2023habitat3}.

Given the importance of virtual indoor scenes to the above applications, computational design tools which ease their creation would be valuable.
Generative models, i.e. systems which can sample novel scenes from a distribution of interest, are a particularly promising technology for this purpose.
Such models can be used to suggest possible placements for new objects in a scene~\cite{zhou2019scenegraphnet}, suggest completions for partial scene designs~\cite{FastSynthCVPR}, or even synthesize entirely new scenes from whole cloth~\cite{Paschalidou2021NEURIPS, tang2023diffuscene, Gao2023scenehgn}.
These capabilities can be used to build tools for interactive design or for the automated creation of large-scale virtual worlds.

The prevailing methodology for building generative models of indoor scenes is to train machine learning models on datasets of existing 3D room layouts.
It is time-consuming and expensive to produce such datasets at the scale required by modern machine learning methods; as such, only a handful of these datasets exist~\cite{fu20213d, Yadav_2023_CVPR}.
These existing datasets contain room labeled from a finite set of room types (e.g. bedrooms, living rooms, offices), and each room is populated with objects from a curated set of 3D object models belonging to a small finite set of object categories (e.g. tables, chairs, beds).
We can think of generative models trained on these datasets as being \emph{closed-universe}: they know up-front the small, finite set of object and room types that they will ever have to produce.

Could one create an \emph{open-universe} generative model which can synthesize any type of indoor scene containing whatever types of objects are needed by that scene?
It has only recently become possible to contemplate this question with the development of so-called ``foundation models'': large machine learning models pre-trained on enormous datasets of text and/or images~\cite{openai2023gpt4, clip, stable_diffusion, mizrahi2023m}.
For example, prior work has shown how to use pre-trained text-to-image generative models to synthesize 3D content which satisfies an arbitrary text prompt~\cite{hoellein2023text2room, jain2021dreamfields, poole2022dreamfusion, yi2023gaussiandreamer}.
While these systems are compelling, they produce output in the form of a single unstructured mesh or density/radiance field; these representations frequently exhibit artifacts and do not easily support editing a scene by moving, swapping, or deleting objects.
In contrast, most prior methods for indoor scene synthesis produce a layout of individual objects, each of which is represented by a high-quality 3D mesh retrieved from a database.

In this paper, we present an open-universe generative model of 3D indoor scenes which produces such structured object layouts in response to a text prompt (Fig.~\ref{fig:teaser}). 
Building such a system requires solving several subproblems.
First, the system must determine the objects that should be in the scene and where they should be located.
To solve this problem, we leverage the commonsense world knowledge encoded by a pre-trained large language model (LLM)~\cite{openai2023gpt4}.
Empirically, we find that tasking an LLM with directly specifying the locations of scene objects leads to poor performance, likely due to the mismatch between metric location coordinates and the vast majority of natural language text contained in its training corpus.
Instead, we task the LLM with producing a \emph{declarative program} in a domain-specific language that describes the objects in the scene and a variety of spatial relation constraints between them.
Executing these programs produces a constraint satisfaction problem which the system solves using a gradient-based optimizer to find one or more object layouts which satisfy the specified constraints.

Once the layout of objects is determined, the system must next insert a 3D mesh for each object in the layout.
One could consider using text-to-3D generative models to synthesize these meshes~\cite{jun2023shape, nichol2022pointe, poole2022dreamfusion}, but as mentioned above, these models can exhibit artifacts and do not (yet) produce outputs with comparable quality to human-created 3D meshes.
Thus, like prior work on indoor scene synthesis, our system retrieves 3D meshes from a 3D object database.
However, unlike prior work, our object retrieval system is designed for the open-universe setting: retrieving from million-scale databases of unlabeled, inconsistently-oriented 3D meshes~\cite{objaverse, objaverseXL}.
We develop a ranking and filtering algorithm using a combination of pre-trained models~\cite{openai2023gpt4, Zhai_2023_ICCV} to retrieve a 3D mesh which matches the attributes of an object specified in the object layout.
We also leverage these models to automatically determine the front-facing direction of each retrieved object, allowing the system to correctly orient each retrieved object as specified in the layout.

We evaluate our system by using it to generate a large variety of different types of rooms, ranging from common indoor spaces (e.g. ``a bedroom'') to rooms designed for specific activities (e.g. ``a musician's practice room'') to outlandish/fantastical scenes (e.g. ``a wizard's lair''). 
For generating typical indoor rooms, we compare to prior methods for closed-universe scene synthesis which are trained on existing 3D scene datasets.
Our system produces scenes which are preferred to those generated by these prior methods in a forced-choice perceptual study.
For open-universe scene synthesis, we compare to LayoutGPT, a recently-published method for generating layouts using large language models. 
Our system also outperforms it in forced choice perceptual study.
We also conduct ablation studies on each component of our system to validate our design decisions.

In summary, our contributions are:
\begin{enumerate}[(i)]
    \denselist
    \item A DSL for specifying indoor scene layouts through declarative constraints and a gradient-based executor for this DSL capable of realizing a distribution of valid scenes from a single program
    \item A robust prompting workflow that leverages LLMs to synthesize programs in our DSL from a high-level natural language description of a scene
    \item A pipeline using pretrained vision-language models for retrieving and orienting 3D meshes from a large, unannotated database to fit object specifications from a scene program
    \item Protocols for evaluating open-universe indoor synthesis systems, including a benchmark set of input descriptions covering a wide variety of possible rooms.
\end{enumerate}

Our code will be made available as open source upon publication.

%% file: 02-related.tex
\section{Related Work}

\paragraph*{Indoor Scene Synthesis}
Indoor scene synthesis has been a problem of interest in computer graphics for years.
In an interesting instance of history repeating itself, some of the earliest work in this area formulated the problem as text-to-scene generation, albeit via laboriously hand-crafted rules~\cite{WordsEye}.
Later, researchers built systems for laying out objects to be consistent with a set of manually-defined design principles~\cite{ifurniture_design}, simple statistical relationships between objects extracted from a small set of examples~\cite{yu2011MakeItHome}, or with programmatically-specified constraints~\cite{rj_mcmc}.

After this, indoor scene synthesis research focused on data-driven methods, using a variety of machine learning methods: Bayesian networks and Gaussian mixture models~\cite{fisher2012}, factor graphs~\cite{kermani2016learning}, topic models~\cite{liang2017}, and stochastic grammars~\cite{qi2018human}.
Once deep neural networks gained popularity, a wave of research applying them to indoor scene synthesis followed: method were proposed using convolutional networks~\cite{FastSynthCVPR, DeepSynthSIGGRAPH2018, wang2019planit}, tree and graph neural networks~\cite{GRAINS, wang2019planit, zhou2019scenegraphnet}, generative adversarial networks~\cite{zhang2019hybrid}, transformers~\cite{Paschalidou2021NEURIPS, wang2020sceneformer}, and finally denoising diffusion models~\cite{tang2023diffuscene}.
All of these prior works present closed-universe generative models, and all of them require (in some cases quite large) datasets of 3D scenes for training.

Recently, the development of pre-trained large language models (LLMs) has raised the possibility of a new generation of text-to-scene generative models, more flexible and open-ended than the early systems from decades ago.
LayoutGPT~\cite{feng2023layoutgpt} is an LLM-based system designed for generating image layouts in an a CSS-like format; the authors also show applications to indoor scene synthesis, albeit for the closed-universe case.
In work concurrent to ours, the Holodeck system shows the ability to use LLMs to generate environments for training embodied AI agents~\cite{yang2023holodeck}.
This system resembles ours in some aspects, including supporting general text prompts instead of fixed room types and specifying object locations implicitly via relations.
It also differs from ours in significant ways: using a simpler specification for object relations (we use a DSL embedded in Python); lacking mechanisms for automatically correcting errors in LLM output; solving for object layouts on a grid, rather than continuously (so objects cannot be adjacent to one another).
Most importantly, it retrieves objects from a curated set of annotated and aligned 3D models, so it cannot be considered truly open-universe.

\paragraph*{Open-vocabulary Text-to-3D}
There has been a recent surge in work leveraging pre-trained vision-language models~\cite{clip, stable_diffusion} to produce 3D content from arbitrary text prompts without any training data.
The most prevalent type of such system works via ``optimization-based inference,'' optimizing for a new 3D output in response to each new text prompt~\cite{jain2021dreamfields, poole2022dreamfusion, yi2023gaussiandreamer, lin2023magic3d, Chen_2023_ICCV, wang2023prolificdreamer}.
Another line of work seeks to amortize this inference procedure by training feedforward neural networks to produce 3D output from a distribution of text inputs~\cite{Sanghi_2022_CVPR, sanghi2023clipsculptor, nichol2022pointe, jun2023shape, lorraine2023att3d}.
The outputs of these methods are either point clouds, unstructured meshes, or isosurfaces extracted from density fields, which are (to date) lower-quality than human-created 3D models.
Additionally, since these systems leverage models trained on images to synthesize 3D structures, they can also suffer from multiview inconsistency artefacts, such as the infamous ``Janus face'' issue~\cite{poole2022dreamfusion}.
Their output also cannot be easily modified, because it is not decomposed into individual objects.

These systems are designed with single-object generation in mind but have been extended to open-vocabulary scene synthesis. 
By combining 2D generative image out-painting with a depth alignment module, Text2Room~\cite{hoellein2023text2room} generates textured meshes of 3D rooms for a given text prompt. Other works~\cite{fang2023ctrl, schult23controlroom3d, bai2023componerf, gao2023graphdreamer} allow the specification of a 3D semantic object layout, which is then used along with text-to-image models for generating textured meshes of scenes/rooms. 
These method suffer from the same mesh quality drawbacks as their single-object counterparts
They also assume an object layout as input, whereas our method generates one.
Our approach could potentially be combined with methods in this space; for example, generative re-texturing of retrieved 3D meshes to fit a desired style or theme~\cite{aladdin}.

\paragraph*{3D Shape Analysis with Foundation Models}
In addition to using pre-trained vision language models (VLMs) for 3D content generation, researchers have also explored how to use these models to analyze existing 3D content without requiring 3D supervision.
Methods have been proposed for captioning/annotating 3D objects~\cite{cap3d, yuan2023cadtalk}, segmenting 3D shapes into semantic parts or identifying regions of interest~\cite{
liu2023partslip, zhou2023partslip, abdelreheem2023SATR, decatur2022highlighter}, and even establishing correspondences between 3D shapes~\cite{abdelreheem2023zeroshot}
Our system uses VLMs to retrieve shapes from a large database, determine their category, and determine their front-facing orientation.

\paragraph*{Program Synthesis with Large Language Models}
One of the key components of our system is using an LLM to generate a declarative program which specifies the layout of objects in a scene. Other work has also explored the use of LLMs to generate programs. Multiple works~\cite{alphacode, alphacode_2} explore the use of LLMs for competitive programming, demonstrating the prowess of LLMs (augmented with symbolic search) at synthesizing programs for a given natural language task description. LLMs' program synthesis abilities have also been employed for solving complex geometric reasoning problems~\cite{AlphaGeometryTrinh2024} and discovering novel mathematical concepts~\cite{FunSearch2023}. Beyond generating programs, some recent works also explore the use of LLMs to improve domain-specific languages automatically~\cite{grand2023learning}. 

\begin{figure*}[t!]
    \centering
    \includegraphics[width=\linewidth]{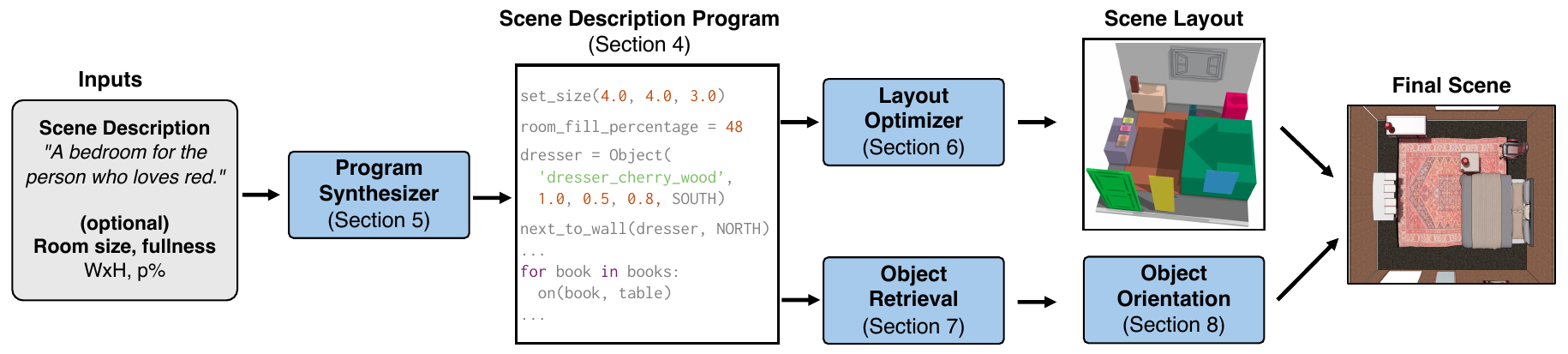}
    \caption{
    A schematic overview of our system.
    Given a high-level natural language description of a scene (plus optional constraints on the room size and object density), an LLM-based program synthesizer produces a scene description program which specifies the objects in the scene and their spatial relations.
    Our layout optimizer module then solves the constraint satisfaction problem implied by this program to produce a concrete layout of objects in the scene.
    For each scene object, the object retrieval module finds an appropriate 3D mesh from a large, unannotated mesh database; the object orientation module then identifies its front-facing direction so that it can be correctly inserted into the scene.
    }
    \label{fig:overview}
\end{figure*}

Recently, systems have been proposed for visual question answering (VQA) by using LLMs to generate programs in a domain-specific language (DSL) designed for image reasoning and then executing that program to answer a given question~\cite{surismenon2023vipergpt, visprog}. 
LLMs have also been used for structured image synthesis by equipping them with a DSL which helps specify 2D object layouts~\cite{Cho2023VPT2I}. Equipped with a library of Python functions in Blender, LLMs have also been used to synthesize 3D scenes, albeit for a closed set of scenes and objects supported by the library~\cite{sun20233dgpt}.

%% file: 03-overview.tex
\section{Overview}
\label{sec:overview}

We aim to solve the following problem: given a natural language description of a desired 3D room-scale scene, produce a 3D scene composed of positioned and oriented 3D meshes retrieved from a database such that the output scene satisfies the input description.
The input description can be flexible: it could provide detailed instructions about the contents of the scene (e.g. ``an office with two desks, a potted plant, and a sofa'') or be intentionally nebulous (``a serious business office'').
For additional control, we also support optional inputs in the form of desired dimensions (in meters) for the output room and how full the room should be (in terms of percentage of floor area occupied by objects).
On the output side, we assume that the generated room has four walls, and that each object is oriented to face along one of the four cardinal directions (north, south, east, west).
These assumptions are reflective of many, but not all, real-world rooms; in Section~\ref{sec:conclusion}, we discuss ideas for removing these assumptions.

Figure~\ref{fig:overview} shows a schematic overview of our system.
Our system specifies the objects which should be in the generated scene and their spatial layout via a scene description program written in a declarative domain-specific language embedded in Python (Section~\ref{sec:language}).
To produce this program, the system feeds the inputs to a program synthesizer module (Section~\ref{sec:progsynth}); internally, this module uses a series of calls to an LLM to write the scene description program.
The complete scene description program is then passed to the layout optimizer (Section~\ref{sec:layoutopt}), which converts the program into a constraint satisfaction problem which it then solves using a gradient-based optimization scheme, producing locations and orientations for all objects.
In addition, the object declarations in the scene description program are sent to an object retrieval module (Section~\ref{sec:retrieval}), which retrieves from a large, unannotated database the 3D model which best matches the description of each object.
Finally, the object orientation module (Section~\ref{sec:orientation}) determines the front-facing direction of each retrieved object, allowing them to be inserted in the room according to the optimized layout to produce the final output scene.

%% file: 04-language.tex
\section{Describing Scenes with Programs}
\label{sec:language}

\begin{figure*}[t!]
    \centering
    \includegraphics[width=\linewidth]{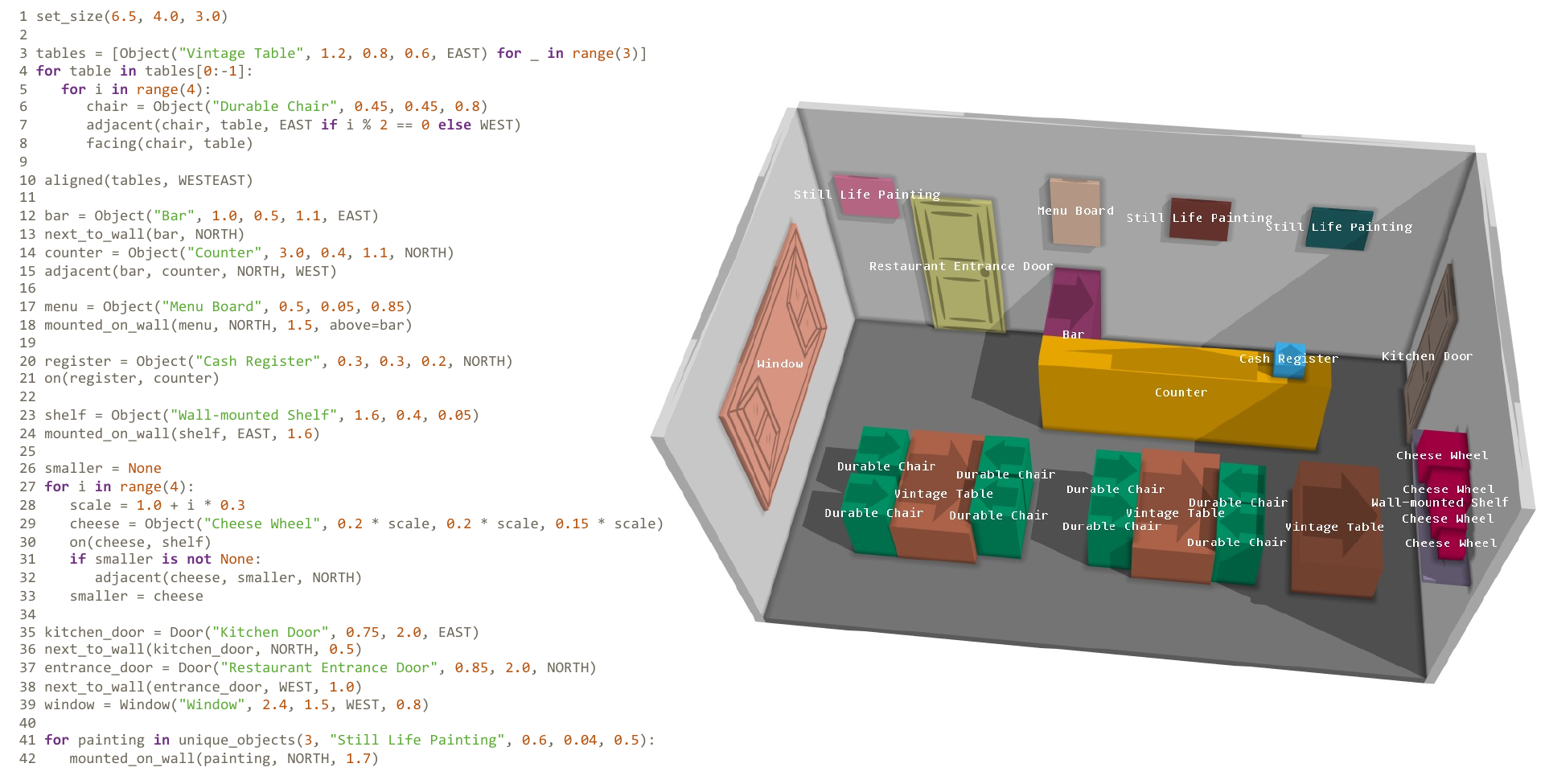}
    \caption{
    An example program in our declarative scene description language (left) and the object layout produced by running this program through our layout optimizer (right).
    This scene depicts a small, cozy Italian restaurant.
    }
    \label{fig:program_example}
\end{figure*}

In this section we introduce our declarative domain-specific language (DSL) for scene description. The simplest way to describe a scene is to explicitly specify all object positions and orientations. However, recent work has shown that even the state of the art LLMs such as OpenAI's GPT4 struggle with accurate placement of objects and object parts~\cite{makatura2023large}. For example, when asked to specify explicit coordinates, GPT4 often creates overlapping objects and objects that are floating in the air. Thus, instead of specifying explicit coordinates, we describe scenes in a declarative manner using spatial relations. Our intuition is that it is easier for an LLM to reason about sentences such as ``the lamp is on the table'' or ``the chair is adjacent to the table'' than about precise numeric values.

Our scene description language is embedded in Python.
Using Python allows us to capitalize on the expressivity of modern programming languages: features such as loops, conditionals, arithmetic, list comprehensions, and many useful built-in functions.
It also takes advantage of GPT4's strong Python programming abilities, likely due to the large amount of Python code in its training corpus.
Our language adds several domain-specific functions to Python. The new functions are either (1) object constructors, (2) relation functions, or (3) parameter setting functions.
Appendix~\ref{sec:apndx_language} contains a full listing of the new functions added by our language.

Figure~\ref{fig:program_example} shows an example program written in our language and the scene layout that it produces when run through our layout optimizer module (Section~\ref{sec:layoutopt}).
This program describes a small Italian restaurant.
Below, we walk through the functionality of each part of this program:
\begin{itemize}
\item Line 1 sets the size of the scene in meters.
\item Lines 3--8 create tables and chairs in a double loop. Most objects are created with an \texttt{Object(description, width, depth, height, facing)} constructor. The width of an object is a dimension perpendicular to the object's front-facing direction; the depth is a dimension along this direction. Relation \texttt{adjacent} constrains chairs to be next to their corresponding tables (two chairs on each side of the table). Relation \texttt{facing} orients each chair to face its corresponding table.
\item Line 10: relation \texttt{aligned} constrains all table centers to be on the same line running west to east.
\item Lines 12--15 create an L-shaped configuration of a bar and counter. Relation \texttt{adjacent(bar, counter, NORTH, WEST)} constrains the bar to be adjacent to the counter from the north and aligned with the west side of the counter.
\item Lines 17--18 create a menu board and constrain it to be mounted on the north wall above the bar, to be accessible both by customers and by staff.
\item Lines 23--33 create a shelf mounted on the east wall and constrain a row of cheese wheels of various sizes to be on top of this shelf.
\item Lines 35--39 create doors and a window. We constrain the kitchen door to be no more than 0.5m from north wall, and the entrance door to be no more than 1m from the west wall.
\item Lines 41--42 create 3 paintings mounted on the north wall. However, unlike chairs, tables and cheese wheels, we don't want the paintings to be represented by the same 3d model. That is why we use a \texttt{unique\_objects} constructor.
\end{itemize}

%% file: 05-progsynth.tex
\section{Generating Scene Programs}
\label{sec:progsynth}

\begin{figure*}[t!]
    \centering
    \includegraphics[width=\linewidth]{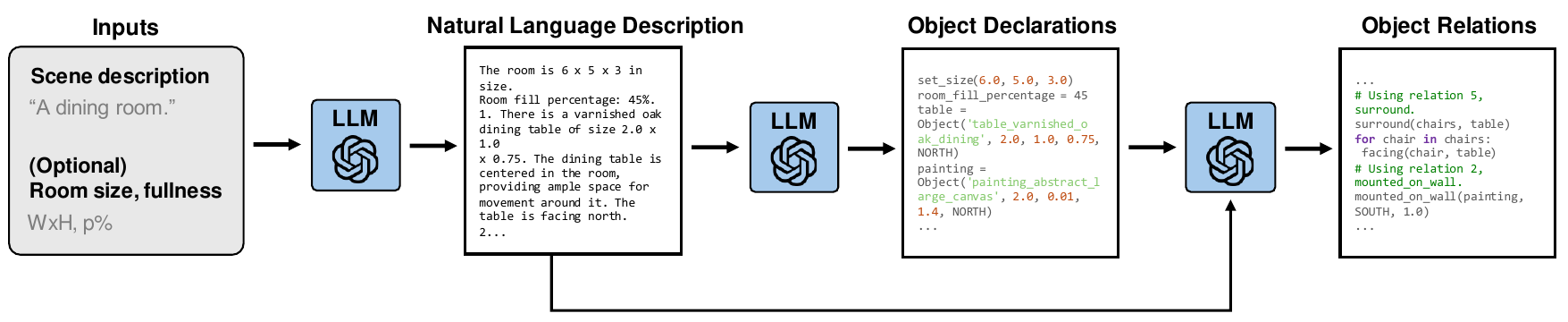}
    \caption{
    Our scene description program synthesizer proceeds in three steps, each of which uses a large language model.
    First, the LLM is asked to generate a natural language description of all the objects in the scene, along with how and why they are spatially related to one another.
    Then, a sequence of two LLMs translate this description into code which declares objects and relations, respectively.
    }
    \label{fig:progsynth_pipeline}
\end{figure*}

Given freeform text input, our system must synthesize scene programs like the one in Fig.~\ref{fig:program_example}.
We use a large language model (LLM) to perform this task.
Going from a high-level, natural language description to a Python program is a challenging task.
LLMs have a remarkable ability to perform this task, but they are not perfect.
In our early experiments tasking an LLM to generate scene description programs directly from text prompts, it tended to produce undesirably short/sparse programs and make some errors (we taxonomize the types of these errors in Section~\ref{sec:error_correction}).
To reduce the prevalence of errors and produce richer scene description programs, we found it helpful to split the program generation task into a series of smaller sub-tasks (an instance of chain-of-thought prompting~\cite{wei2023chainofthought}):
\begin{enumerate}
    \item Generate a detailed natural language description of the complete scene
    \item Generate the code to declare all objects in the scene (ensuring there are enough objects to achieve the desired room fullness)
    \item Generate the code to specify all object relations
\end{enumerate}
Task (1) is better aligned with the LLM's training data, so it is more reliable than directly generating code.
Tasks (2) and (3) become easier because they can refer to a concrete natural language description---essentially, they are translation tasks, rather than synthesis tasks.
Figure~\ref{fig:progsynth_pipeline} shows a schematic of this pipeline.
The remainder of this section describes each of these stages in more detail; the complete prompt templates for each can be found in the supplemental material.
In Section~\ref{sec:results}, we conduct an ablation study to show the value of this task decomposition.

\subsection{Describing the Scene in Natural Language}

The first stage of the program synthesizer tasks an LLM with describing the scene to be generated in natural language.
First, if the user has not provided an input room size or target object density, the LLM is first asked to produce values for those quantities which are appropriate for the input text prompt.
The texture of the walls and floors are described as well, to be retrieved later in the pipeline.
Then, the LLM is asked to list and thoroughly describe all the objects in the scene: the type of object, its size, and any salient details about its appearance.
For each object, the LLM also outputs a description of how that object is situated in the scene in relation to other objects.
Throughout, the LLM is asked to explain its reasoning.

To guide the LLM, our prompt template for this stage includes two in-context examples of the type of output we expect in this stage. 
The first in-context example describes an artist's one room apartment; the second describes a typical a dining room.
These two examples span a variety of object types and arrangements; our results in Section~\ref{sec:results} show that the system generalizes beyond these two examples to synthesize an even wider variety of scenes.

\subsection{Declaring Objects}

The next stage of the program synthesizer tasks an LLM with producing Python code that declares all the objects in the scene.
This stage receives the natural language description output by the first stage as its input.
In addition to choosing the most appropriate constructor for each object (i.e. \texttt{Object}, \texttt{objects}, or \texttt{unique\_objects}) and the relevant arguments for those constructors (description, dimensions, and facing direction information), it also produces its estimate of a `category label' for the object (e.g. for `a sleek dark wood dining table,' the category label might be `table'). 
This category label is used by the later object retrieval and orientation stages of the pipeline.

The prompt template for this stage of the program synthesizer also includes two in-context examples to help the LLM produce code in the correct output format.
These in-context examples show object declaration code for the same two scenes described in the first stage's in-context examples (the artist's apartment and the dining room).

\paragraph*{Achieving target object density}
Specifying the target occupied room floor area in the input ishelpful for encouraging the LLM to generate the right amount of objects, but it does not guarantee that the LLM will produce output that satisfies this target.
Thus, this stage includes logic that checks if the target occupied area has been achieved by the sizes of the generated object declarations; if not, it invokes another LLM to generate more objects (both their natural language descriptions and their object declaration code).
This step is iteratively repeated until the actual percentage of occupied room floor area meets the set target object density.

\subsection{Specifying Object Relations}
The final stage of the program synthesis pipeline receives all previous pipeline outputs and completes the scene description program by generating code describing object relations.
This task boils down to translating the free-text descriptions of object arrangement in the Stage 1 output into calls to appropriate relation functions in our DSL (which refer to the appropriate objects declared by Stage 2).

The prompt template for this stage also includes two in-context examples.
These in-context examples contain relation specification code based on the same two scenes that the previous two stages used as in-context examples.
We designed these in-context examples to demonstrate certain potentially non-obvious ways to use relations to achieve layout goals (e.g. using two \texttt{next\_to\_wall} relations to place an object in a corner).

%% file: 06-layoutopt.tex
\section{Scene Layout Optimization}
\label{sec:layoutopt}
In this section, we describe how a scene program is converted into an object layout, i.e. a list of objects with locations and orientations. This process consists of two stages. First, the Python interpreter converts the program into a geometric constraint satisfaction problem, where variables are object positions and object directions and constraints are derived from relation functions. Second, the constraint problem is solved using an algorithm based on gradient descent.
Because LLMs are not perfect programmers, scene programs can contain errors. In the last part of this section, we describe how the system handles different types of errors.

\subsection{Converting Scene Programs into Constraint Problems}
First, the Python scene program is executed with a Python interpreter. As described in Section~\ref{sec:language}, the new functions in the scene DSL include object constructors and relation functions. Standard object constructors define variables of the constraint problem. Relation functions define constraints of the constraint problem. Door and window constructors define both variables and constraints.

While we designed the scene DSL to be most convenient for LLMs to use, we designed the constraint set to be most simple mathematically. Each relation function within the DSL is translated into one or more of ten defined constraints: \texttt{ON}, \texttt{NEXTTOWALL}, \texttt{HEIGHT}, \texttt{ADJACENT0}, \texttt{ADJACENT1}, \texttt{ADJACENT2}, \texttt{CEILING}, \texttt{ABOVE}, \texttt{ALIGNED} and \texttt{FACING}. This translation is mostly straightforward. For example, \texttt{mounted\_on\_wall(a, wall, height\_above\_ground, above)} is translated into 
\begin{verbatim}
NEXTTOWALL(a, wall, 0.0),
HEIGHT(a, height_above_ground),
ABOVE(a, above, wall).
\end{verbatim}
The non-obvious cases are relation functions that take list arguments: \texttt{aligned} and \texttt{surround}. Relation function \texttt{aligned(list, direction)} is desugared into a list of \texttt{len(list)-1} constraints:
\begin{verbatim}
ALIGNED(list[0], list[1], direction)
ALIGNED(list[1], list[2], direction)
...
\end{verbatim}
Relation functions \texttt{surround(objects, centerobj)} are handled last. Surrounding objects in the \texttt{objects} list are processed one-by-one. Each \texttt{object} is constrained to be adjacent to the \texttt{centerobj} from the side of the \texttt{centerobj} that has the most free space available, and facing the \texttt{centerobj}. This process respects other adjacencies and walls.

In addition to the relational constraints, we also add default constraints: \texttt{WITHINBOUNDS} and \texttt{NOOVERLAP}. For every object \texttt{a} we add \texttt{WITHINBOUNDS(a)} to ensure that the object stays within the bounds of the room.
For every unordered pair of distinct objects \texttt{a}, \texttt{b} we add \texttt{NOOVERLAP(a, b, distance\_x, distance\_z)}. Here \texttt{distance\_x} and \texttt{distance\_z} are zeros if neither of objects \texttt{a}, \texttt{b} is a door or a window, and some nonzero parameter if \texttt{a} or \texttt{b} is door or a window. The meaning of these distance arguments is to create 'auras' for doors and windows that no other objects can overlap. This is needed to ensure that doors can be opened, and windows are not obstructed by furniture. There is one exception to this rule: we allow wide flat objects such as rugs to overlap with anything, to support the 
case (common in real furniture arrangements) where a part of a furniture object stands on a rug.

\subsection{Solving the Constraint Problem}
The purpose of the layout optimizer is to find a vector of object positions and object directions that satisfies all the constraints. Given the geometric nature of our constraints, it is natural to formulate our constraint solving problem as an optimization of a (mostly) differentiable function. The only non-differentiable constraint that we have is a \texttt{FACING} constraint (since orientations are restricted to the four cardinal directions); we address this constraint separately.

We design differentiable loss functions for each constraint (excluding \texttt{FACING}), such that a constraint is satisfied if and only if its loss is zero. For example, \texttt{HEIGHT(a, height)} is the squared difference between object's \texttt{a} bottom vertical coordinate and the \texttt{height} parameter, and \texttt{WITHINBOUNDS(a)} is the sum of squares of object's \texttt{a} linear extensions beyond the scene cuboid. We refer the reader to Appendix~\ref{sec:apndx_optimizer} for the full list of constraint losses.
Our system finds a solution to a constraint problem by initializing objects in random positions and minimizing the sum of constraint losses with gradient descent. Since the initial configuration is random, different runs of optimizer can produce different layouts (see Figure~\ref{fig:multi_opt}).

\paragraph{Custom gradients for non-overlap constraints}
For constraints of the form \texttt{NOOVERLAP(a,b)}, the natural loss formulation is a measure of the overlap of the bounding boxes of \texttt{a} and \texttt{b}, and this is indeed what we use. However, such functions are flat if one cuboid is inside the other along each axis (for example, if two cuboids overlap in a shape similar to the $+$ sign, as in the inset figure).
\begin{wrapfigure}{r}{0.3\linewidth}
    \includegraphics[width=\linewidth]{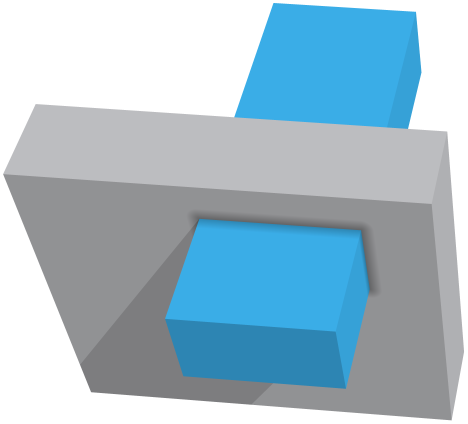}
\end{wrapfigure}
Hence, the gradient can be zero when the loss is not zero, and gradient descent fails to minimize the loss.
To solve this issue, we define the gradient of the \texttt{NOOVERLAP(a,b)} constraint to be proportional to the vector connecting the centroids of \texttt{a} and \texttt{b}, where the magnitude of this vector is equal to the minimum side length of the cuboidal overlap region between \texttt{a} and \texttt{b}.

\begin{figure}[t!]
    \centering
    \setlength{\tabcolsep}{3pt}
    \begin{tabular}{cc}
         Without Repel Forces & With Repel Forces \\
         \includegraphics[width=0.48\linewidth]{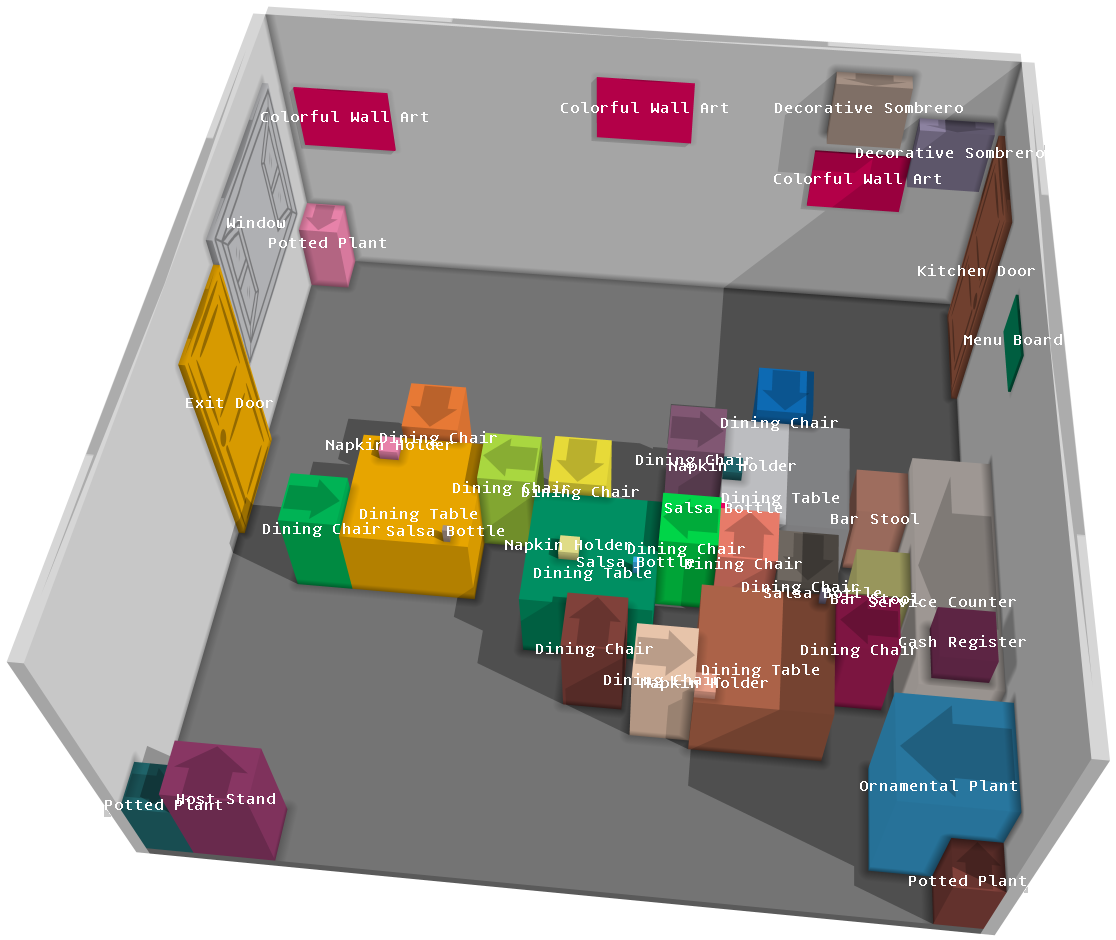} &
         \includegraphics[width=0.48\linewidth]{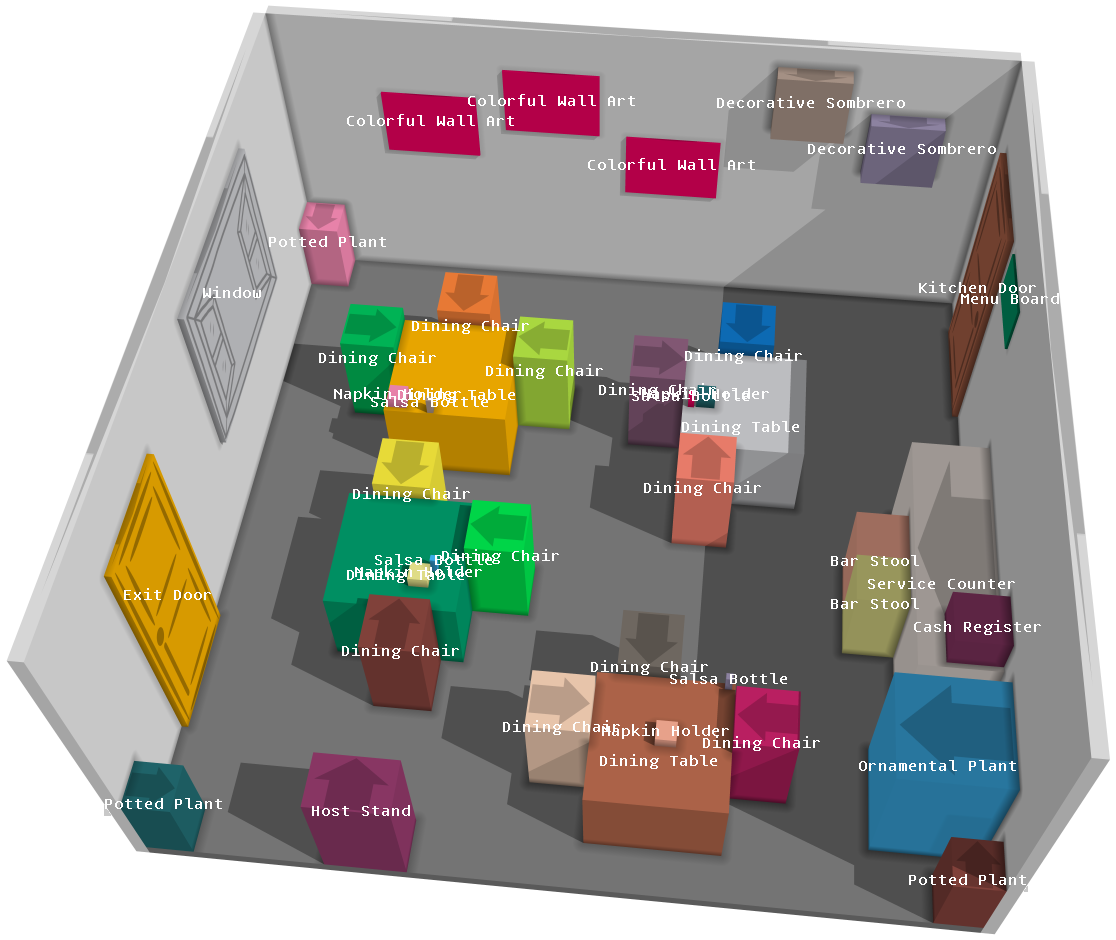}
    \end{tabular}
    \caption{Adding repel forces to layout optimization allows objects to be appropriately spaced without exhaustively specifying explicit relations.}
    \label{fig:layout_ablation}
\end{figure}

\paragraph{Repel forces}
\label{par:repels}
Most scene description programs in our language are underspecifications of scenes: there exist many object layouts which can satisfy the specified constraints.
Are there any priors we can use to inform whether any of these layouts are better than others?
One reasonable assumption is that since object adjacencies have such a strong perceptual impact on the scene (causing objects to be perceived as part of some larger group), the optimized layout should not have adjacencies that are not explicitly specified in the program.
We realize this assumption using repel forces, similar to repels in force-directed graph drawing~\cite{GraphDrawingBook} (implementation details in Appendix~\ref{sec:apndx_optimizer}).
Repel forces make the distribution of objects in a scene more balanced but do not push away objects that should be together. This makes a notable difference in the plausibility of optimized scene layouts; see Figure \ref{fig:layout_ablation}. However, gradient descent with repel vectors added to the gradient does not necessarily converge to a solution of the original constraint problem. We describe the solution to this issue in the next paragraph.

\paragraph{Determine object orientations}
In the scene description language, an object can be specified to face either one of four cardinal directions or another object. If an object faces some cardinal direction, its direction is known up-front and does not change. Situations where one object faces another (for example, where a sofa faces a TV) are trickier. To know the final direction, the system must know final object positions. However, we cannot set directions after position optimization: if we rotate an object, its bounding box would change, which could make the object overlap with other objects or violate other constraints. Thus, object directions should be optimized jointly with object positions.
This complicates our gradient descent scheme, because direction is a discrete variable.

We developed a simple, multi-stage optimization approach that solves this problem and also gets around the convergence issue with repel forces mentioned at the end of the previous paragraph:
\begin{enumerate}
\item Randomly initialize object positions and non-fixed object directions
\item Perform gradient descent with repel forces added to the gradient
\item Set non-fixed objects directions according to the current object positions
\item Perform gradient descent again without repel forces
\end{enumerate}
Object positions do change in the second descent phase, so in the end, some object directions may not agree with object positions. However, the second descent phase usually moves objects only slightly, and the mismatch between object positions and directions happens very rarely in practice.

\subsection{Error Correction}
\label{sec:error_correction}
As mentioned in Section~\ref{sec:progsynth}, LLMs tasked with synthesizing scene description programs can sometimes produce code with errors.
Our decomposition of the program synthesis task into stages helps reduce these errors (as we show in Section~\ref{sec:results}), but it does not completely eliminate them.
Since these errors typically affect only a small part of an otherwise-valid program, the layout optimizer includes mechanisms for automatically fixing some of these errors to avoid throwing out the entire LLM-generated program.

We taxonomize the types of errors the program synthesizer makes into four classes:
\begin{itemize}
    \item \textbf{Hallucination:} calling functions which are not in our language or referring to objects which do not exist in the scene (e.g. \texttt{below(footrest, desk)}, or \texttt{on(lamp, table)} when \texttt{lamp} does not exist)
    \item \textbf{Misuse:} incorrectly using a function in the language (e.g. incorrect argument type, missing arguments)
    \item \textbf{Contradiction:} creating relations which are provably in direct conflict with one another (e.g. \texttt{next\_to\_wall(statue, NORTH)} and \texttt{next\_to\_wall(statue, SOUTH)})
    \item \textbf{Unsatisfiability:} creating a set of relations which do not have any obvious conflicts but which cannot be jointly satisfied (i.e. the layout optimizer converges to non-zero loss).
\end{itemize}

The layout optimizer catches the first two types of errors by running the Python interpreter, catching any raised exceptions, deleting the line which caused the exception, and trying to execute the program again.
We classify an exception as a hallucination if it contains the string \texttt{``is not defined''} and as a misuse otherwise.

After the program is successfully executed, the layout optimizer identifies contradiction-type errors by searching for the following patterns (identifiable as subgraphs in the overall constraint graph):
\begin{enumerate}
\item An object is adjacent to, stands on top of, or faces itself.
\item An object is next to two opposing walls and is not large enough to span the room dimension.
\item Object \texttt{a} is adjacent to object \texttt{b} from direction \texttt{d}, and also \texttt{b} is adjacent to \texttt{a} from any direction other than the opposite of \texttt{d}.
\item Object \texttt{a} stands on top of object \texttt{b}, and also \texttt{b} is (horizontally) adjacent to \texttt{a}.
\item Object \texttt{a} is next to wall \texttt{d}, but some other object is adjacent to \texttt{a} from direction \texttt{d}.
\item The total length of objects adjacent to \texttt{a} from some direction is more than the corresponding linear size of \texttt{a}.
\end{enumerate}
The layout optimizer handles detected contradictory subgraphs by deleting the constraint in the subgraph which is declared last in the scene program.

If the system does not reach a near zero loss after 10 rounds of layout optimization from random initial conditions, we consider the constraint problem to be unsatisfiable (the fourth type of error).
In this case, the system backtracks to the third stage of the program synthesizer and re-generates the relations part of the scene description program.

%% file: 07-retrievealign.tex
\section{Retrieving 3D Objects}
\label{sec:retrieval}

\begin{figure*}[t!]
    \centering
    \includegraphics[width=\linewidth]{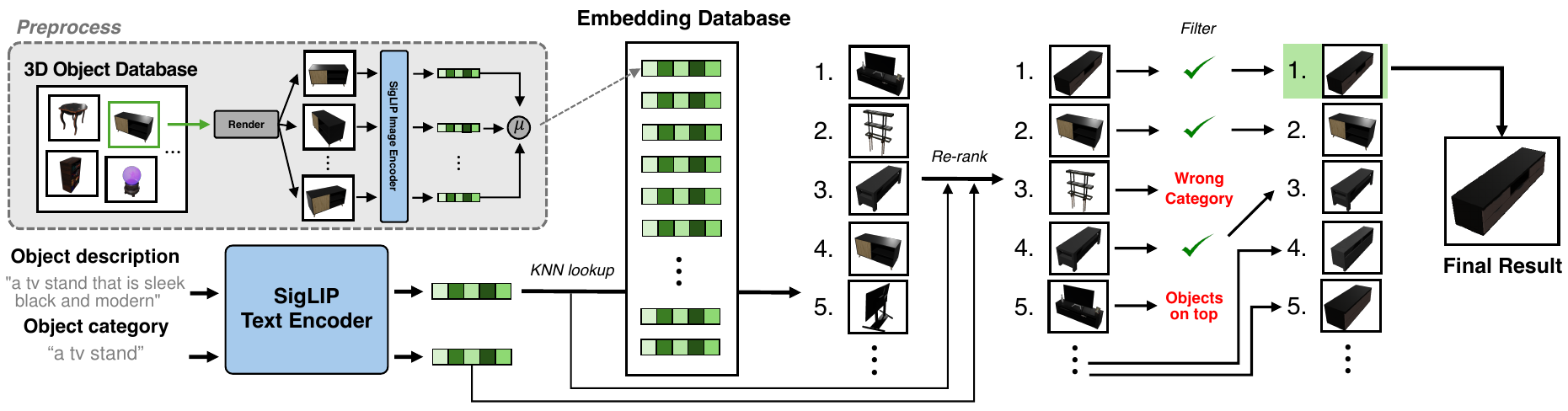}
    \caption{
    Our pipeline for open-universe 3D object retrieval.
    As a preprocess, we compute embeddings for each object in our 3D mesh database using a vision language model (VLM).
    Given a description and category of an object (both specified in the LLM-generated scene program), our system finds the $k$ nearest neighbors of the text description's VLM embedding in our database.
    These initial retrieval results are then re-ranked to prioritize objects with the correct category and further filtered to remove meshes which are the wrong category or which consist of multiple objects.
    }
    \label{fig:retrieval}
\end{figure*}

Given a generated scene description program, the system must retrieve relevant 3D object meshes for each object declared in the program. In this section, we describe our object retrieval pipeline specifically designed for retrieving objects from massive, un-annotated 3D asset datasets.
Figure~\ref{fig:retrieval} shows a schematic overview of this stage.

Given an object's natural language description $T$ (automatically-generated by the LLM-based program synthesizer), our goal is to retrieve a list of relevant 3D meshes from our 3D asset dataset $\mathcal{D}$. As we want to match a textual description $T$ to a dataset of 3D assets, a natural approach is to leverage large pretrained Vision Language Models (VLMs) such as CLIP~\cite{clip}. VLMs enable measuring similarity between the text $T$ and renderings of the 3D meshes by generating their embeddings and measuring their cosine similarities. Therefore, we can retrieve a list of $k$ relevant 3D meshes $\mathcal{D}^T_k$ for the given object description $T$ as follows:
\begin{equation}
\mathcal{D}^T_k = \text{TOP}_k(f_t(T), \mathcal{D}_E)
\label{eq:retrieval}
\end{equation}
where $f_t$ is the text encoder of the VLM and $\text{TOP}_k(x, Y)$ returns entries $y \in Y$ with the top-k highest cosine similarities to $x$. $\mathcal{D}_E  = \{ E(x) | x \in \mathcal{D} \}$ is a database of VLM embeddings, one per object in $\mathcal{D}$.
The per-object embedding function $E(x)$ is defined as rendering 12 views of the object $x$, computing the VLM image embedding of each, and averaging those embeddings.
We create this embedding database as a preprocess.
Despite the massive scale of the database, performing this retrieval step takes only $0.5$ seconds, as we employ approximate nearest neighbor search using FAISS~\cite{FAISS}. The memory requirements of building $\mathcal{D}_E$ are also limited, as we only need to store a single embedding vector for each 3D asset.

While this retrieval algorithm is a good starting point, we found it to be insufficient in practice. Due to the large, unstructured, and noisy nature of massive 3D asset datasets, we observe and correct for some frequently-occurring failure cases, which we describe below.

\subsection{Retrieving the Correct Category}
We observe that often the retrieved 3D mesh might be visually be similar to the text prompt but functionally be of a different category. As the text prompt contains information about both the object category (e.g. chair, table etc.) and the object style (e.g. wooden, modern etc.), retrieved objects can sometimes have a style matching the text description but a different category. To solve this problem, we use the category attribute $C$ that the program synthesizer produces for each object and enhance the retrieval with a \textit{category-aware reranking} strategy. Specifically, we found the following recipe to work well. 
First, we retrieve a large set ($k=100$) of 3D objects from the embedding database $\mathcal{D}_E$, as in Equation~\ref{eq:retrieval}.
Then, we re-rank the retrieved objects $x$ based on a category-aware embedding distance given by
\begin{equation}
    d(f_t(T), E(x)) + \lambda d(f_t(\hat{C}), E(x))   
\label{eq:weighted_rerank}
\end{equation}
where $\hat{C}$ is the text prompt ``A photo of a $C$'', where $C$ is the category attribute, and $\lambda$ is weighting coefficient (we use $\lambda=1$).

For indoor scenes, we have found that retrieving an object of the wrong category can often seriously disrupt the plausibility of the scene.
Therefore, we augment our category aware re-ranking by leveraging a multimodal LLM~\cite{openai2023gpt4} to select an object of the correct category.
Specifically, we provide the LLM with an image of an $n \times n$ array of renders of the top $n \times n$ retrieved objects, tasking it with identifying which images are/are not of the category $C$. This step is applied sequentially on the next $n \times n$ retrievals and so on, until the system finds $m$ objects judged to be of the correct category ($m=1$, except in the case where the program dictates that unique meshes be retrieved for a set of objects).
Note that while one could apply this category filtering directly on the initial retrieved set of objects, applying the category-aware re-reranking first means that fewer filter steps (and therefore fewer costly LLM API calls) are required to find $m$ objects of the correct category.

\subsection{Retrieving Only Single Objects}
Another mode of failure occurs when the retrieved mesh contains additional secondary objects along with the desired object. Objects such as tables and TV stands are often modeled alongside adjacent or supported objects such as chairs and TVs, respectively.
Comparing VLM text embeddings to embeddings of images of such objects often produces high similarities despite the presence of the additional secondary objects. To solve this problem, we again employ a multimodal LLM to perform \textit{multi-object filtering}. Similarly to our category filtering step, we provide the LLM with  a set of object renders and task it with filtering out objects based on two criteria: (i) if there are other objects on top of the main object of category $C$ (e.g. a TV stand with a TV on it) and (ii) if the object has other objects around it (e.g. a table with chairs around it). The step is also applied sequentially on the top $n \times n$ retrievals until we have retained $m$ valid object retrievals.

\begin{figure*}[t!]
    \centering
    \includegraphics[width=\linewidth]{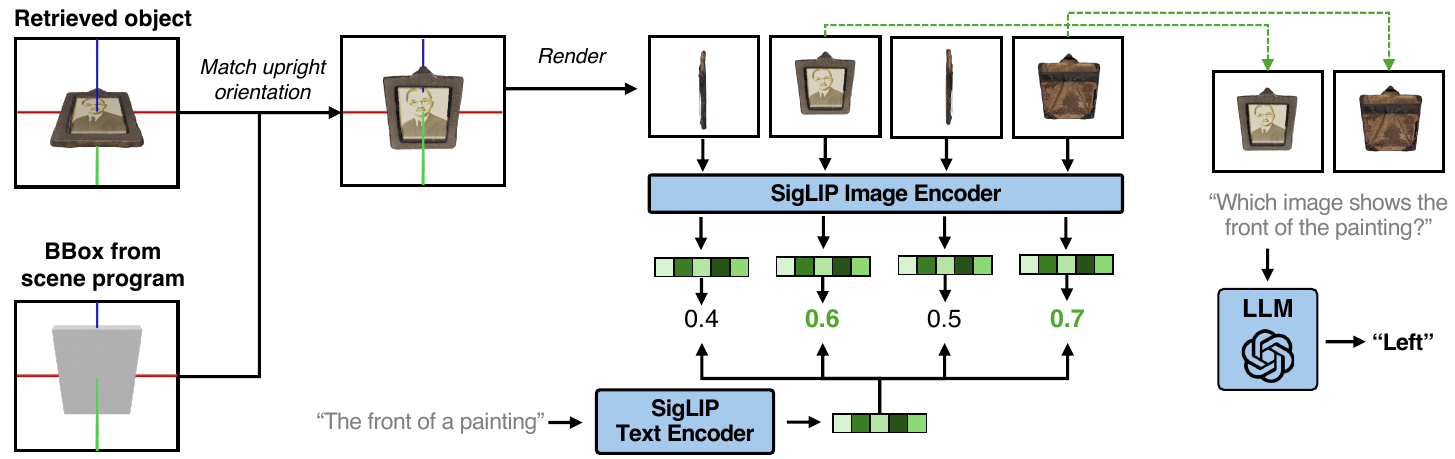}
    \caption{
    Our pipeline for orienting retrieved 3D meshes.
    The system first tries a discrete set of rotations to match the upright orientation of the mesh's bounding box to that of the bounding box specified in the object layout.
    Then, a VLM is used to assess how similar each of four orthogonal views of the mesh are to the phrase ``the front of a $C$,'' where $C$ is the object's estimated category.
    The two most similar views are then given to a multimodal LLM to decide which is the best choice for the front face of the object.
    }
    \label{fig:orientation}
\end{figure*}

\subsection{Matching the Specified Object Size}

Not all 3D meshes which match the object category and description specified in the program are good candidates, because they must also reasonably match the object \emph{size} specified in the program (e.g. a long, thin mesh will not be a good candidate for an object which is specified as being small and squarish).
Thus, we also filter out meshes whose bounding box aspect ratios are too dissimilar from those specified by the object dimensions in the generated program.
Given the bounding boxes $B^r$ of a candidate mesh and $B^p$ of the object as specified in the program, we compute the minimal Bounding Box Distortion (BBD) as follows:
\begin{align}
\label{bbd_metric}
    BBD(B, B_p) &=  \Bigg( \sum_{i \in \{x, y, z\}}{\Big|\log\frac{{B}_i}{B^p_i}\Big|} \Bigg)\\
    mBBD &= \min_{R \in \mathcal{R}} BBD(R \cdot B, B_p)
\end{align}

where $R$ is a rotation from the set of Euler angles $\mathcal{R} = \{$ (0,0,0), (0,90,0), (90,0,0), (90,0,90) $\}$, and $B^*_i$ refers to the volume normalized side length along the axis $i$.
That is, we measure the distortion on whichever of these axis-aligned rotations best aligns $B^r$ with $B^p$.
This small set of candidate rotations is typically sufficient, because artist-created 3D meshes are typically modeled with their up direction aligned with one of the world coordinate axes.
Candidate objects with a minimal distortion larger than a threshold $\tau = 0.4$ are discarded (unless this would retain fewer than $m$ objects, in which case the lowest-loss $m$ objects are kept).

\section{Orienting Retrieved Objects}
\label{sec:orientation}

The final problem we need to solve for end-to-end text to indoor scene generation is correctly orientating the retrieved objects in the scene. Specifically, the system must determine where the `front' of an object faces, which can be critical for objects such as chairs and bookshelves.
Full $SO(3)$ canonicalization of 3D objects is a challenging task, especially without assuming any supervised training data.
Some unsupervised canonicalization methods exist~\cite{sajnani2022_condor}, but they are category-specific; in our case, there are no fixed category labels for our retrieved meshes.
Fortunately, as mentioned in the previous section, most artist-created meshes are modeled with their up direction and front direction aligned with world coordinate axes, allowing us to reduce our problem from a continuous search over $SO(3)$ to a six-way classification problem: which of the six faces of a mesh's axis-aligned bounding box corresponds to its `front.'
In this section, we present a simple, training-free approach for solving this problem using pre-trained models.
Figure~\ref{fig:orientation} shows an overview of our approach.

Our first goal is to ensure that the object is in upright position. Once the object is in upright position, the problem further reduces to detecting which of the four vertical faces of the object corresponds to its `front'. To this end, we use the bounding box distortion (BBD) metric from Equation~\ref{bbd_metric}. 
Empirically, we find that the majority of 3D meshes are modeled in a y-up coordinate system, which our system also uses.
Thus, we only rotate the object if doing so would significantly improve BDD loss w.r.t. the y-up bounding box specified in the scene program.
Specifically, we check if either of the rotations $(90,0,0)$ or $(90,0,90)$ result in a BDD less than the minimum of $(0, 0, 0)$ and $(0, 90, 0)$ by a margin of 1.0; if so, we re-orient the mesh.
At this point, we may be done: if the program synthesizer did not declare the object with a \texttt{facing} argument in the generated scene program, then we assume that the object does not have a unique front-facing direction and does not need to be further oriented (e.g. round tables).

If the object does have the \texttt{facing} argument, we next convert the four-way classification problem into a two-way classification problem.
For objects with a non-square footprint, two of the four options can be directly rejected based on the BBD metric: for example, for rectangular couches, the side faces can be rejected as mapping them to the `front' leads to high BBD loss.
Specifically, if the BBDs for the rotations $(0,0,0)$ and $(0,90,0)$ differ by more than 0.4, we reject the side faces.
For objects with more square footprints, we use a VLM to reject two of the four options.
Specifically, we render the object from the four cardinal directions, embed each of these renders with the VLM's image encoder, and measure the cosine similarity between each of these embeddings and the embedding of the text `The front of a $C$', where $C$ stands for the object category (specified by the LLM).
We discard the directions with the lowest two similarities.
Finally, we solve the remaining two-way classification problem by leveraging a multimodal LLM to perform visual question answering. Specifically, we provide the LLM with the two remaining renders and ask `Which image shows the front of the a $C$?'.

In Section~\ref{sec:results}, we compare this multi-step orientation pipeline with several simpler alternatives (including approaches using only a VLM and only a multimodal LLM), showing that it performs best.

%% file: 08-results.tex
\section{Results and Evaluation}
\label{sec:results}

\begin{figure*}[t!]
    \centering
    \setlength{\tabcolsep}{3pt}
    \newcolumntype{y}{>{\centering\arraybackslash}p{0.24\linewidth}}
    \begin{tabular}{yyyy}
         \textit{``A university dorm''} &
         \textit{``A boss's office''} &
         \textit{``A vampire's room''} &
         \textit{``A medieval knight's room''}
         \\
         \includegraphics[width=\linewidth]{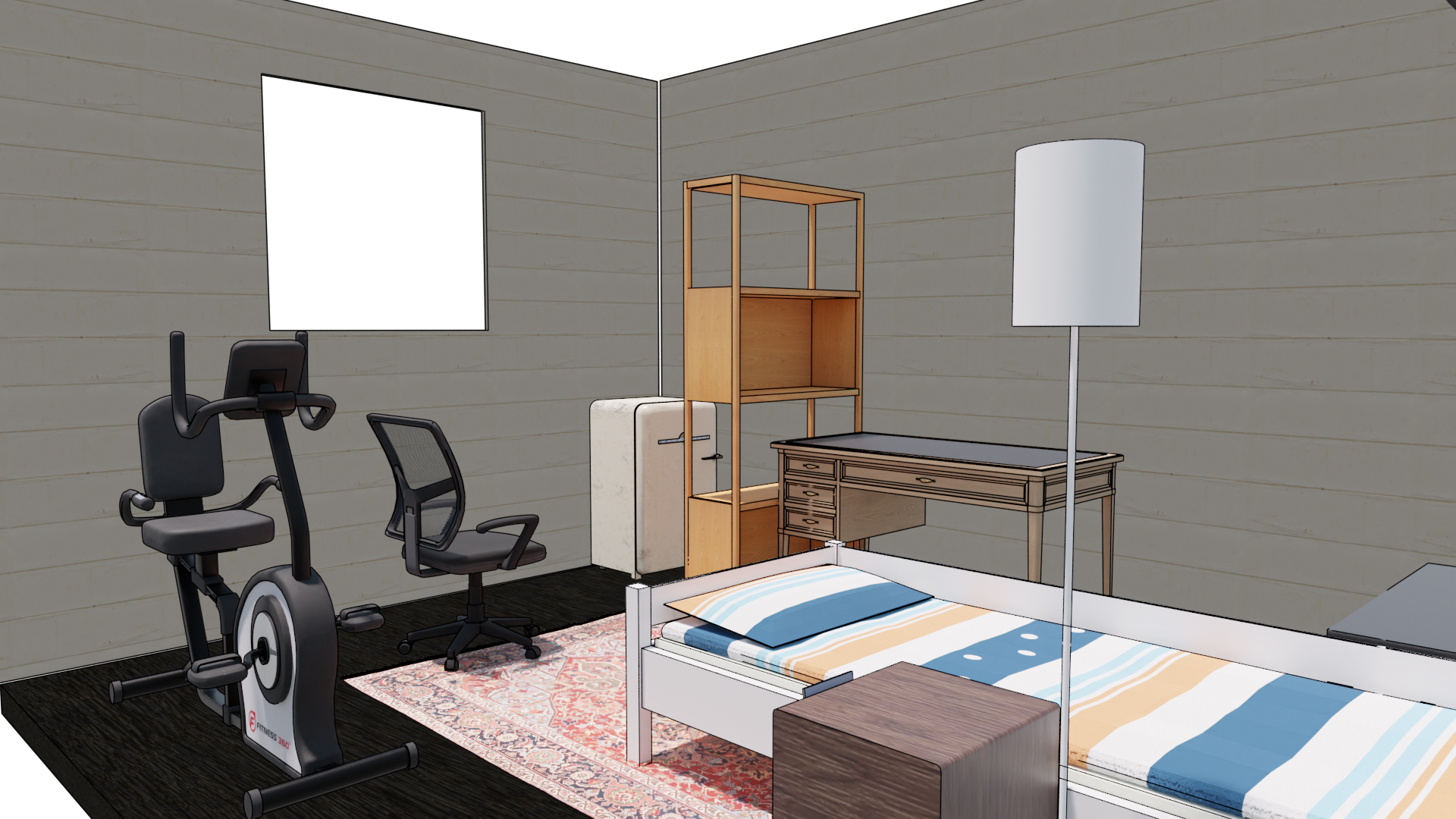} &
         \includegraphics[width=\linewidth]{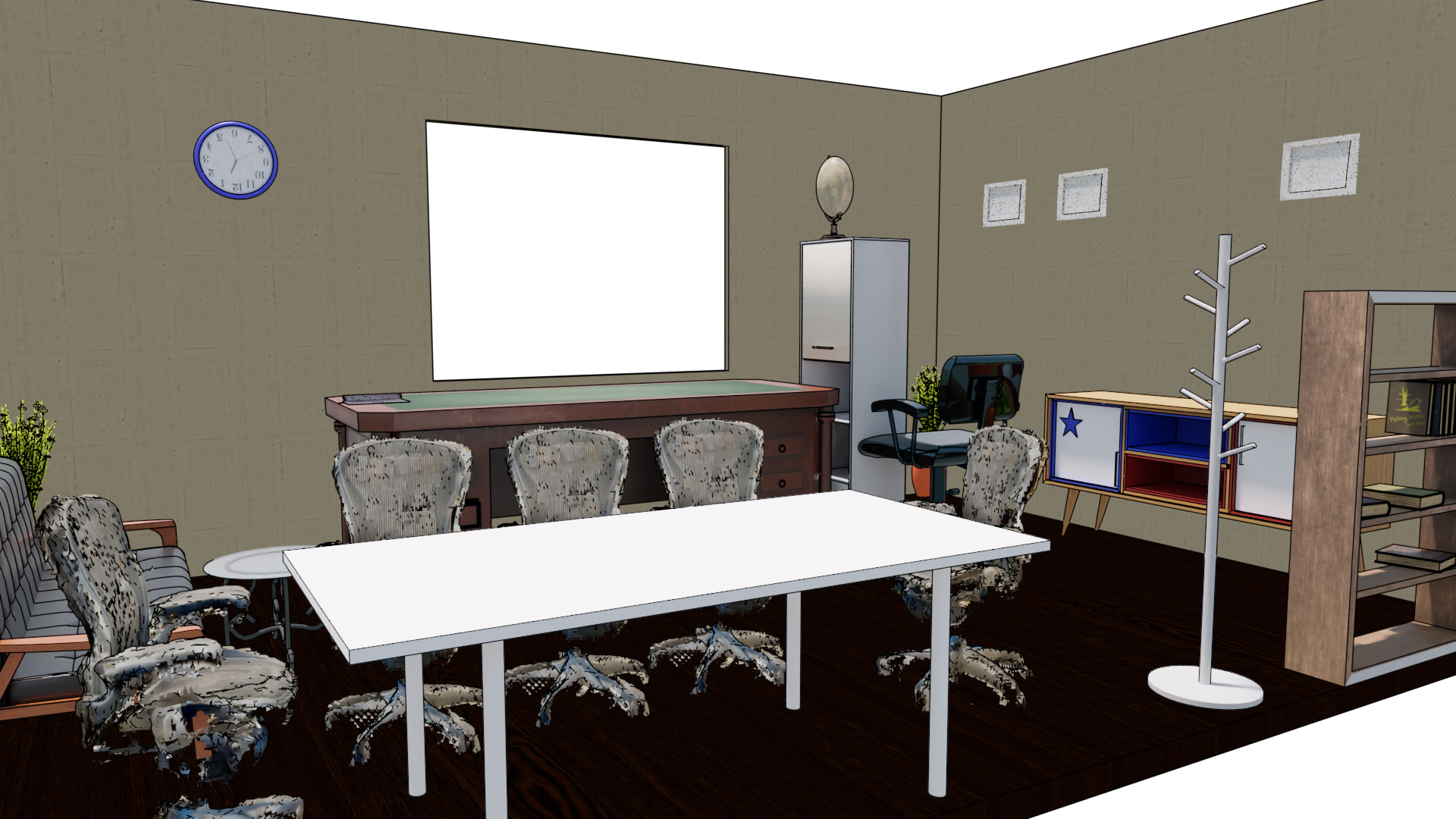} &
         \includegraphics[width=\linewidth]{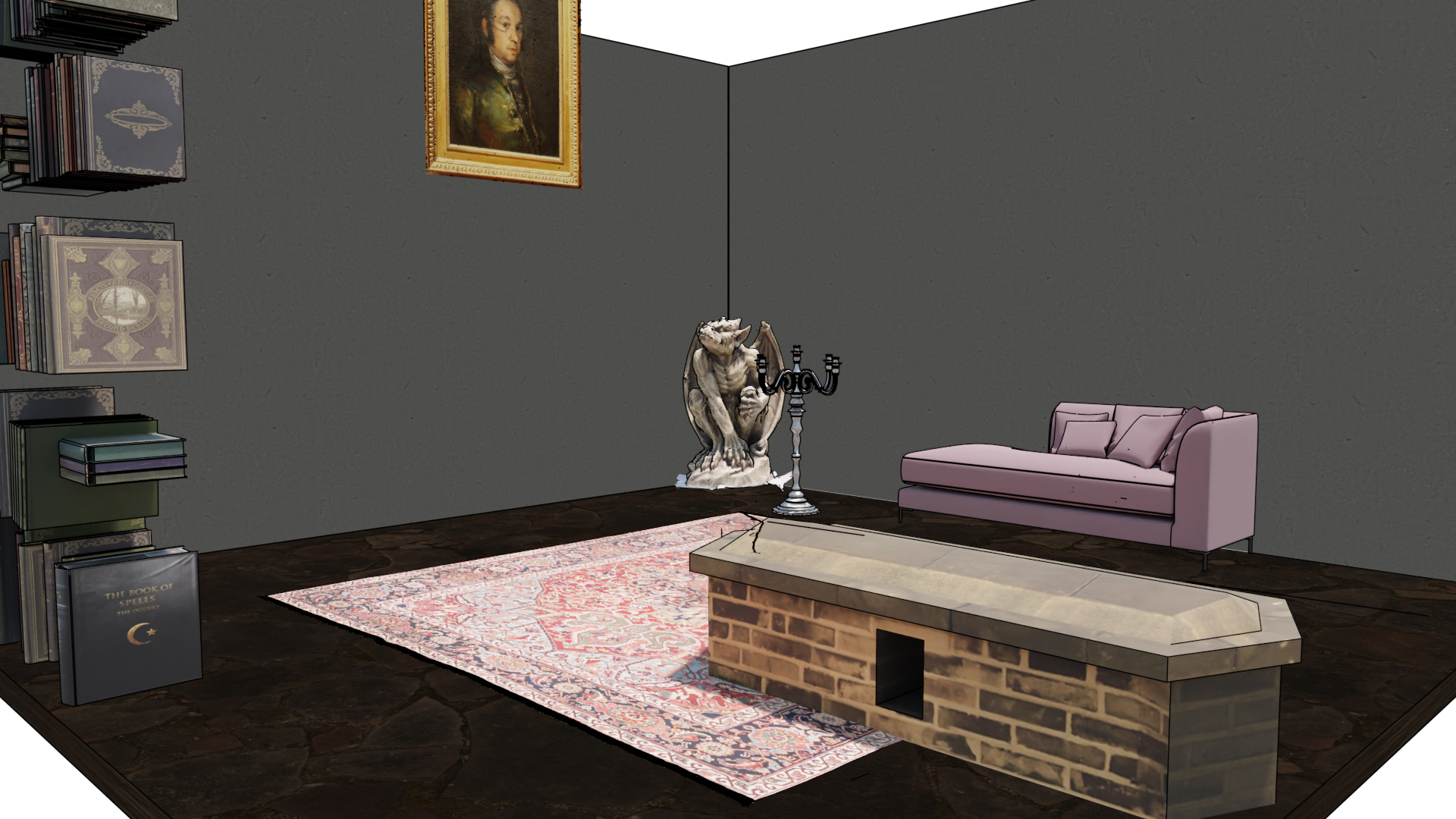} &
         \includegraphics[width=\linewidth]{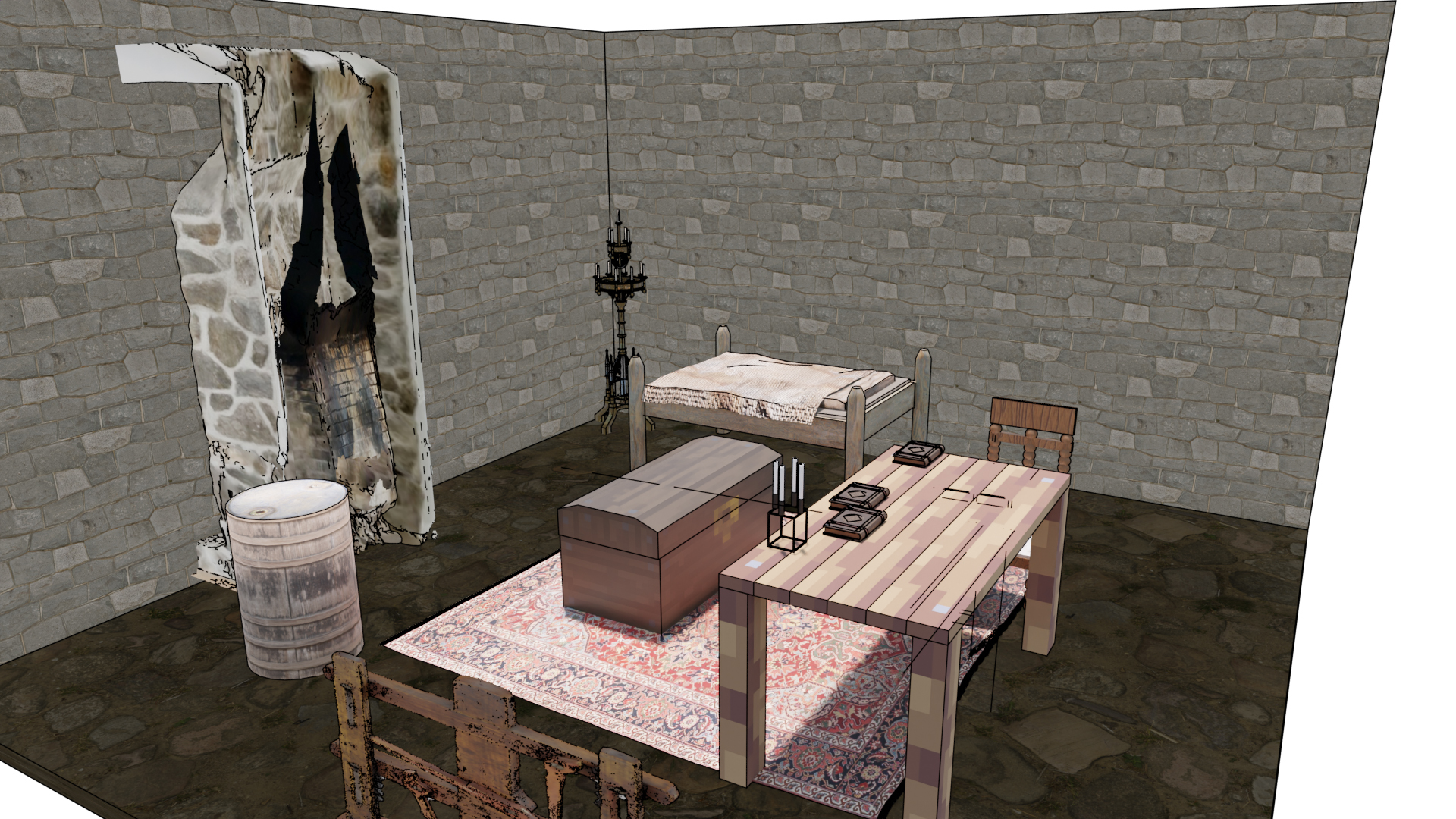}
    \end{tabular}
    \caption{Our method is capable of synthesizing a wide variety of indoor scenes that conform to input text prompts.}
    \label{fig:qual_results}
\end{figure*}

In this section, we evaluate our scene synthesis system by using it to generate scenes in response to a variety of inputs and by comparing it to other scene generation methods, both closed- and open-universe.

\paragraph{Implementation details}
We use GPT4 and GPT4V for all language generation and visual question answering tasks throughout the system~\cite{openai2023gpt4}.
For joint text-image embedding, we use the SigLIP vision-language model~\cite{SigLIP}.
For object retrieval, we use the Objaverse dataset~\cite{objaverse} as well as multi-view renderings of Objaverse objects provided by the ULIP dataset~\cite{ULIP}.

\subsection{Qualitative Results}

\begin{figure}[t!]
    \centering
    \includegraphics[width=\linewidth]{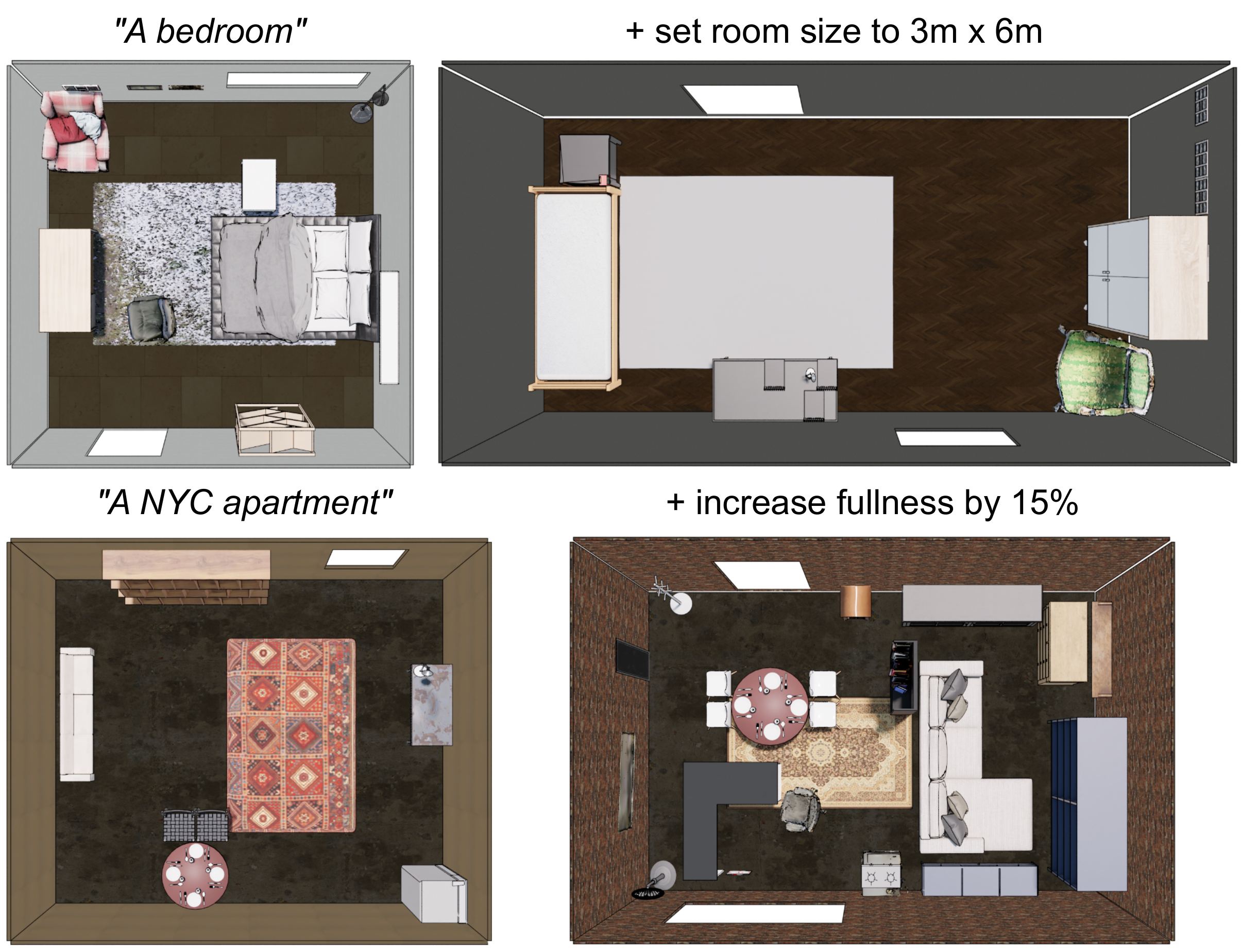}
    \caption{When the user specifies the desired room size or target object density as part of the input, our system appropriately adjusts its output.
    \emph{Top:} controlling room size; \emph{Bottom:} controlling room fullness.}
    \label{fig:user_controls}
\end{figure}

Figs.~\ref{fig:teaser} and \ref{fig:qual_results} show scenes generated by our system in response to different text prompts.
Our system is able to interpret prompts describing typical indoor scenes (e.g. ``a bedroom''), rooms for specific purposes (e.g. ``a dining room for one''), different styles of interiors (e.g. ``an old-fashioned bedroom''), and non-residential indoor spaces (``a high-end mini restaurant'').
It can also imagine scenes that don't exist in the real world (e.g. ``a vampire's room'' with a coffin and a collection of occult tomes; ``a medieval knight's room'').
Our system also supports user specification of the desired room size and object density/fullness; in Fig.~\ref{fig:user_controls}, we show examples of how the system responds to these optional inputs.

\begin{figure}
    \centering
    \setlength{\tabcolsep}{1pt}
    \begin{tabular}{ccc}
         \includegraphics[width=0.32\linewidth]{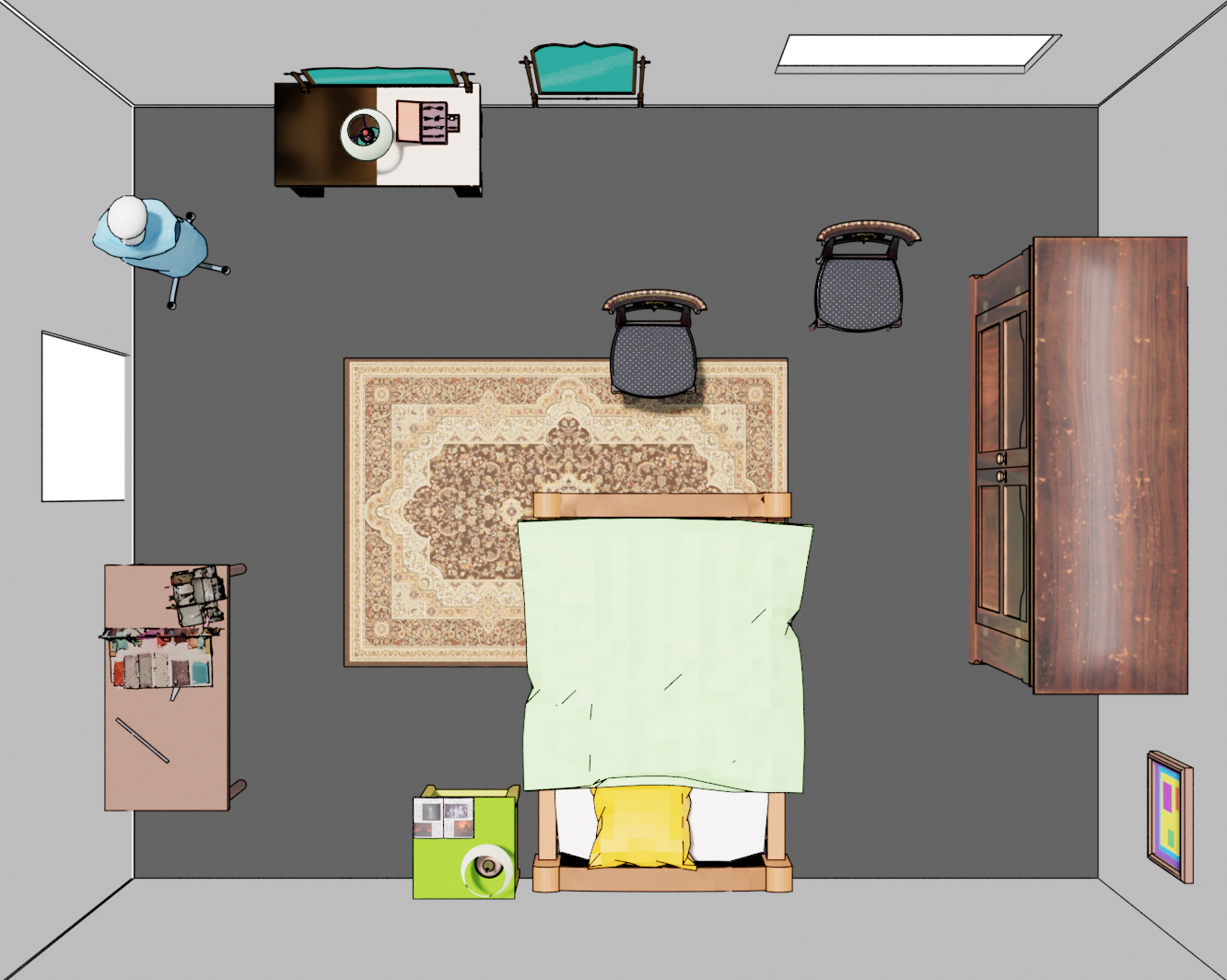} &
         \includegraphics[width=0.32\linewidth]{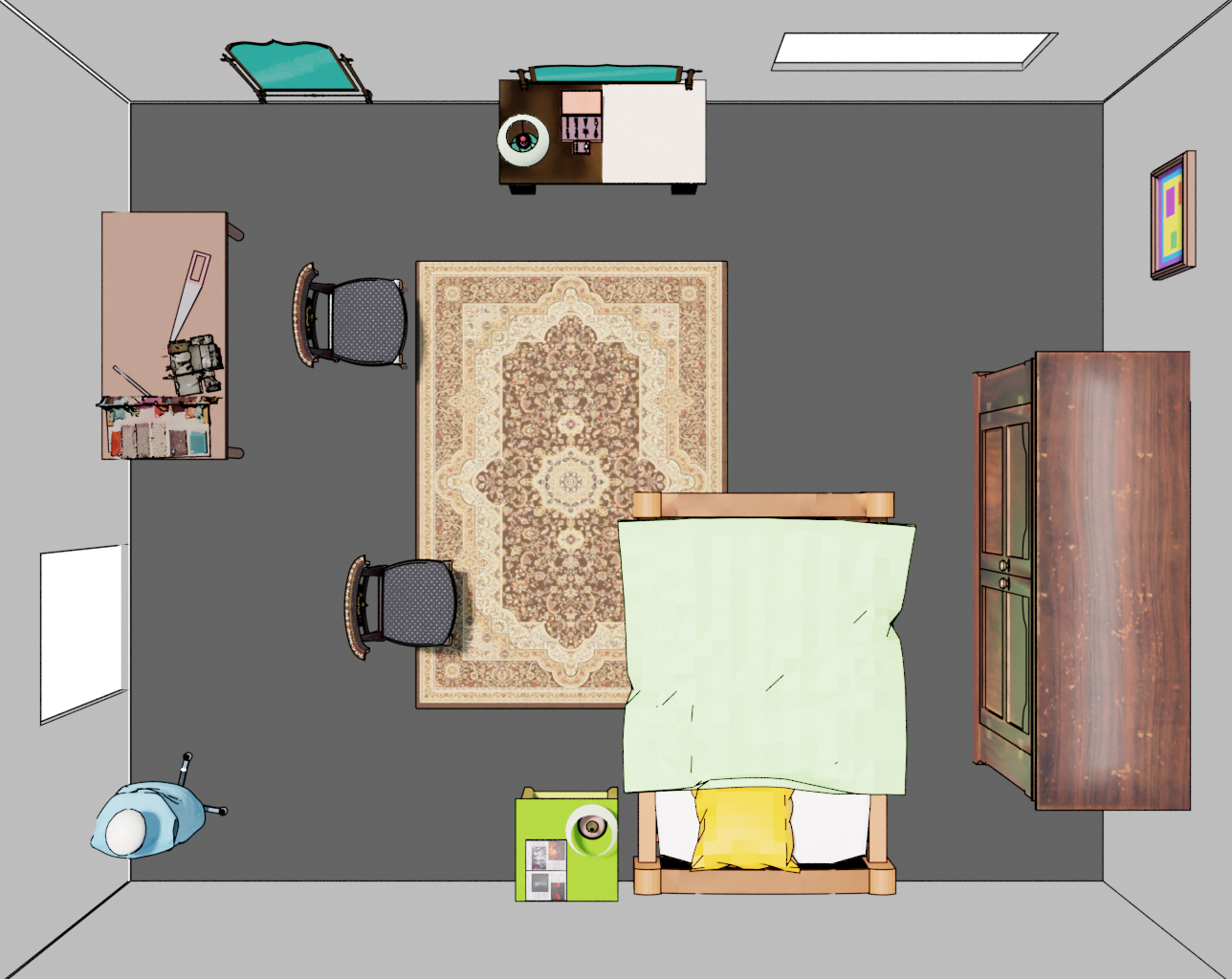} &
         \includegraphics[width=0.32\linewidth]{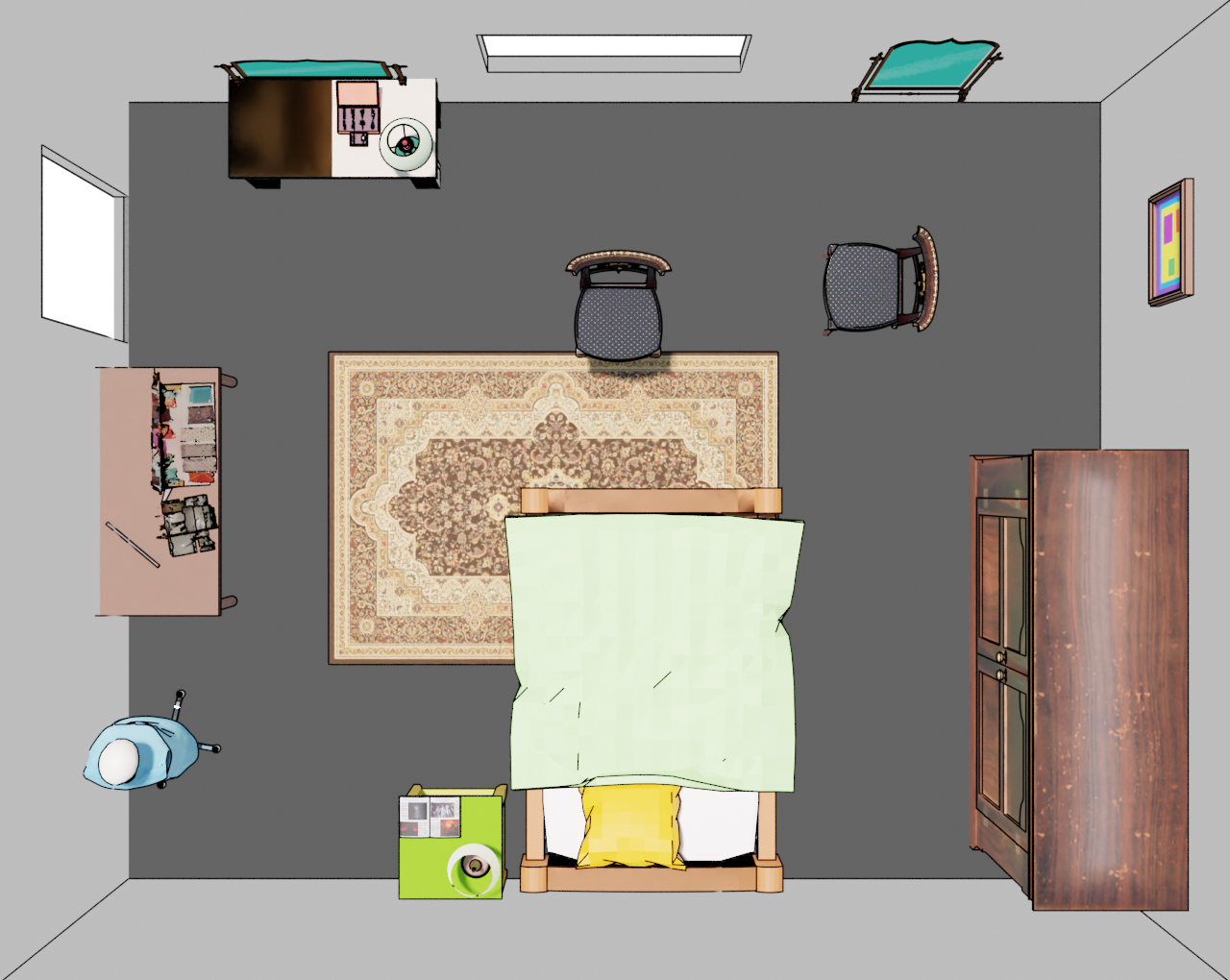}
    \end{tabular}
    \caption{Given a single scene description program, we can run the layout optimizer multiple times to produce stochastic variations on the same scene.}
    \label{fig:multi_opt}
\end{figure}

Because we break the scene synthesis problem into declarative program synthesis followed by layout optimization, it is possible to optimize multiple layouts for the same program, producing multiple design variations.
Fig.~\ref{fig:multi_opt} shows an example of this process.
Objects placed along walls (such as paintings and desks) are free to slide along those walls, in some cases exchanging positions (e.g. the desk can appear on both sides of the door).

\subsection{Closed-Universe Scene Synthesis Comparison}

Here we evaluate how well our system performs on a closed-universe scene generation task when compared to prior methods for this problem which learn from 3D scene data.
Specifically, we compare against the following methods:
\begin{itemize}
    \item \textbf{ATISS}~\cite{Paschalidou2021NEURIPS}: a recent autoregressive Transformer-based generative model of indoor scenes.
    \item \textbf{DiffuScene}~\cite{tang2023diffuscene}: a recent denoising diffusion-based generative model of indoor scenes.
\end{itemize}
We evaluate each of these methods on generating bedrooms, living rooms, and dining rooms (three commonly-occurring room types in closed-universe scene generation work).
We direct our method to generate object layouts of these types by providing it with text prompts of the form ``A bedroom.''

To compare the object layouts generated by these different methods, we conducted a two-alternative forced-choice perceptual study.
We recruited 35 participants from a population of university students.
Each participant was shown a series of 45 comparisons, where each comparison contained a room type label (bedroom, living room, or dining room), images of two scenes, and a question asking them to choose which scene they thought was a more realistic instance of that type of room.

\begin{figure}[t!]
    \centering
    \small
     \includegraphics[width=\linewidth]{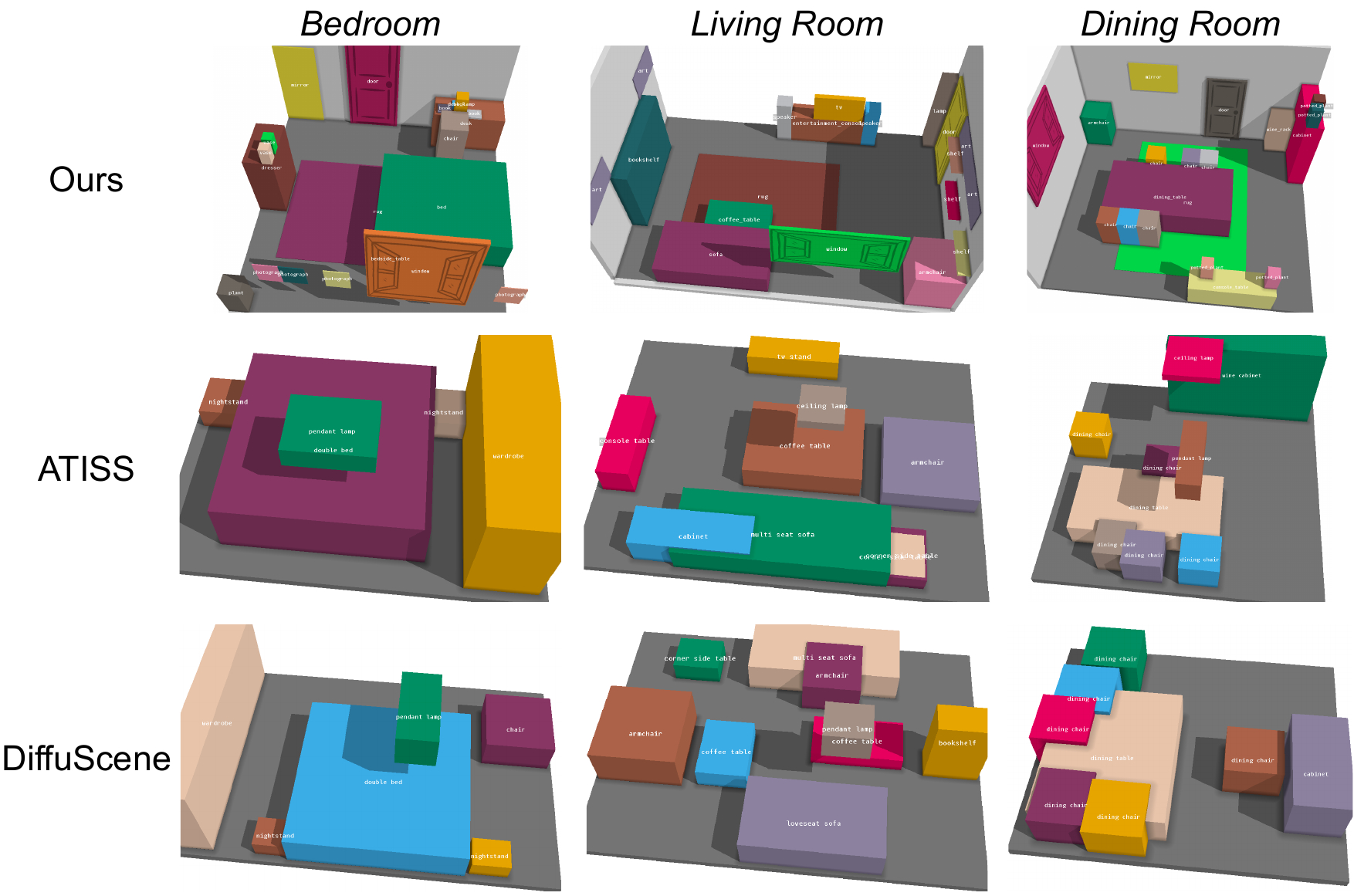}
    \caption{
    Comparing closed-universe scene layouts generated by our system to those generated by two existing closed-universe scene generative models (zoom in to read object text labels).,
    Layouts generated by our method are more detailed and devoid of object intersection artefacts.
    }
    \label{fig:qual_compare_closed}
\end{figure}

Table~\ref{tab:quant_compare_closed} shows the results of this study, and Fig.~\ref{fig:qual_compare_closed} shows some of the object layouts generated by each method.
Participants vastly preferred the object layouts produced by our method compared with ATISS (79\% overall preference rate) and DiffuScene (81\% overall preference rate). 
While we find this trend is consistent across the three room types we used, the gap between our approach and these alternatives is most pronounced for dining rooms, where the objects layout we produced were preferred at rates of 86\% and 89\% over ATISS and DiffuScene respectively.
As seen in the last column of Fig.~\ref{fig:qual_compare_closed}, our method captures relations important to scene fidelity (e.g. surrounding a dining table with chairs), all the while avoiding object overlaps and clutter that mar the scenes produced by the other two approaches. 

\begin{table}[t!]
    \centering
    \begin{tabular}{lcccc}
        \toprule
        \textbf{Ours vs.} & \textbf{Bedroom} & \textbf{Living} & \textbf{Dining} & \textbf{Average} \\
        \midrule
        ATISS & 76\% & 74\% & 86\% & 79\% \\
        DiffuScene & 75\% & 79\% & 89\% & 81\% \\
        \bottomrule
    \end{tabular}
    \caption{
    Results of a two-alternative forced-choice perceptual study comparing scenes generate by our system to those generated by two existing systems for closed-universe scene generation.
    The scenes our method generated were largely preferred over those from alternative approaches across typical indoor room types. 
    }
    \label{tab:quant_compare_closed}
\end{table}

\subsection{Open-Universe Scene Synthesis Comparison}
\label{sec:perceptual_study}

We next evaluate our system's ability to generate open-universe scenes.
To the best of our knowledge, there is no prior work which solves this exact problem.
Thus, we compare against the next best thing: an existing method that uses LLMs for scene synthesis.

Specifically, we compare against LayoutGPT~\cite{feng2023layoutgpt}.
LayoutGPT was originally only evaluated in the closed-universe setting; we adapt it to the open-universe setting by modifying its prompt to remove references to fixed sets of room and object types and by providing it the same in-context examples that our method sees (converted into LayoutGPT's scene representation format).
To convert LayoutGPT's generated layouts into full 3D scenes, we use the same object retrieval and orientation modules as in our system.

We test how well the two methods can generate scenes in response to a range of different types of text prompts, ranging from simple to more complex/subtle: 
\begin{enumerate}
    \item \textbf{Basic:} basic room types such as ``a bedroom.''
    \item \textbf{Completion:} prompts that describe a subset of a basic scene and ask the system to complete it, e.g. ``a living room with a sofa, tv, and a coffee table.''
    \item \textbf{Style:} basic room types with style descriptors, e.g. ``a minimalist living room.''
    \item \textbf{Activity:} rooms that must accommodate some specific activity, e.g. ``a musician's practice room.''
    \item \textbf{Fantastical:} fantastical, outlandish, or whimsical rooms that would not exist in reality, e.g. ``a wizard's lair.''
    \item \textbf{Emotion:} rooms which should evoke specific emotions, e.g. ``A lonely dark jail cell.'' 
\end{enumerate}
We have created 59 prompts across these 6 types; a complete listing of all prompts can be found in the supplemental material.

\begin{table*}[t!]
    \centering
    \begin{tabular}{l c c c c c c c c}
        \toprule
        & \textbf{Basic}
        & \textbf{Completion}
        & \textbf{Style}
        & \textbf{Activity}
        & \textbf{Fantastical}
        & \textbf{Emotion}
        & \textbf{Average}
        \\
        \midrule
        Ours vs. LayoutGPT &
        66\% &
        64\% &
        76\% &
        60\% &
        51\% &
        66\% &
        65\%
        \\
        \bottomrule
    \end{tabular}
    \caption{
    How often open-universe scenes generated by our method are preferred to those generated by LayoutGPT in a two-alternative forced-choice perceptual study (higher is better).
    We report results for the different types of prompts in our evaluation set as well as overall results.
    Our system is preferred over LayoutGPT for all prompt types except Fantastical, where there is no clear preference.
    }
    \label{tab:quant_compare_open}
\end{table*}

\begin{figure*}[t!]
    \centering
    \includegraphics[width=\linewidth,]{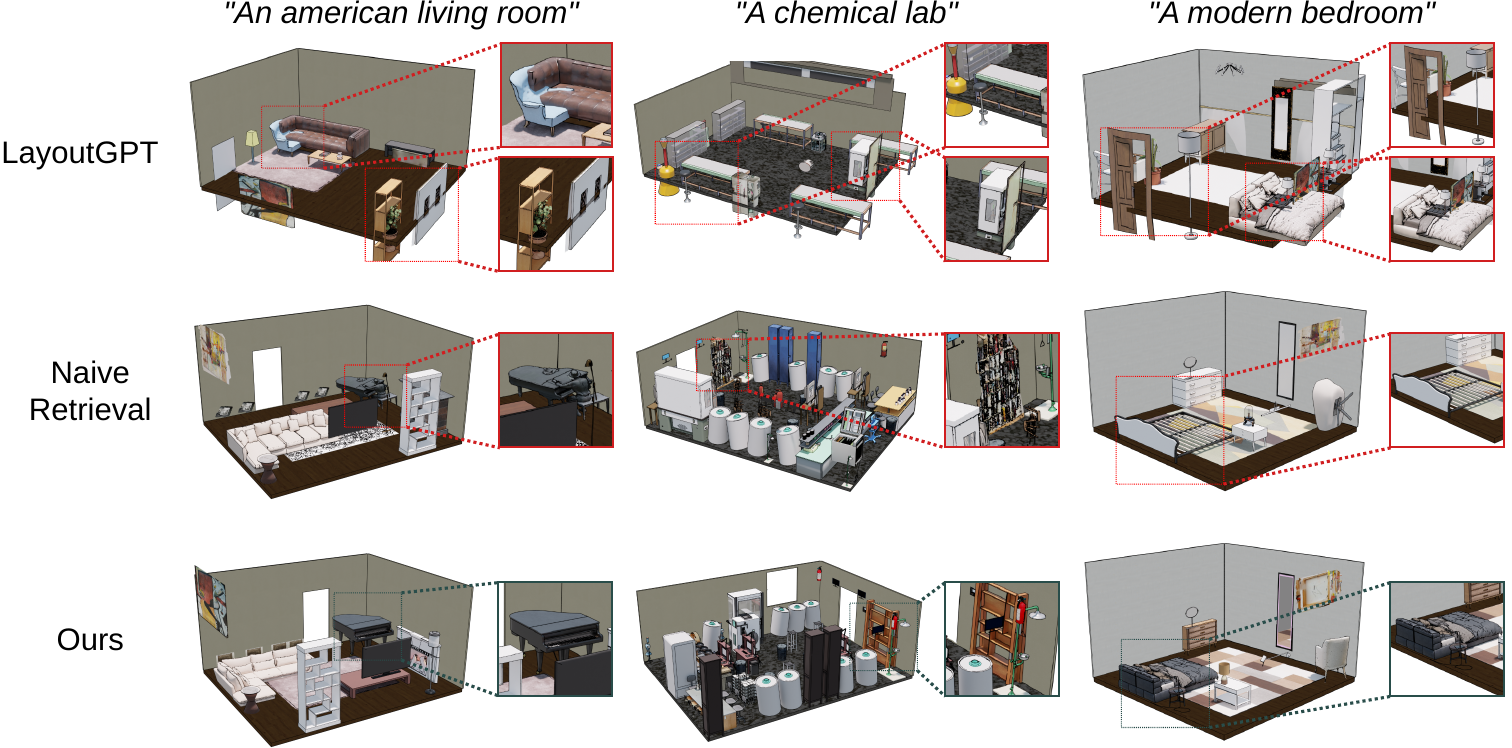}
    \caption{
    Comparing open-universe scenes generated by our system to those generated by LayoutGPT and to scenes generated by an ablation of our system using a naive object retrieval method.
    LayoutGPT produces layouts with many overlapping objects; the naive retrieval baseline sometimes retrieves unusual and undesirable meshes for some objects.
    }
    \label{fig:qual_compare_open}
\end{figure*}

To compare how well the different methods fare on these prompts, we conduct a two-alternative forced-choice perceptual study, pitting our method against LayoutGPT.
We recruited 24 participants from a population of university students.
Each participant was shown a series of 50 comparisons, where each comparison contains a text prompt, images of two scenes, and a question asking them to choose which scene they thought was better (taking into account overall scene plausibility and appropriateness for the prompt).

Table~\ref{tab:quant_compare_open} shows the results of this experiment.
In general, participants preferred our method's scenes over those from LayoutGPT.
The largest margin between our system and LayoutGPT occurred for prompts in the Style category. 
Since both systems used the same object retrieval method, this difference is not attributable to the objects in one condition having more stylistically-appropriate appearance; rather, our system does a better job of interpreting which types of objects should be in a scene and how they should be arranged to satisfy stylistic goals.
The Fantastical prompt category, with its unusual prompts, proved to be challenging for both methods, with no clear winner emerging. 

Fig.~\ref{fig:qual_compare_open} shows some of the scenes generated by each method in this experiment.
To demonstrate the value of our object retrieval module, we also produce variants of scenes generated by our method where the 3D meshes used are retrieved using a naive retrieval method (only the initial VLM-based KNN retrieval step from our full pipeline, without re-ranking or filtering).
LayoutGPT, as it directly generates numerical coordinates for object locations, suffers from frequently interpenetrations between objects.
Our method avoids these errors by construction.
Our full retrieval pipeline also helps avoid some erroneous mesh retrievals (e.g. for the bookshelf in the second column).
The supplemental material contains more examples of objects from these scenes where our full retrieval pipeline retrieves a mesh of the appropriate category but the naive approach does not.
Across all the scenes, for every three out of 100 objects, our method retrieved a correct-category mesh whereas the naive method did not.

\subsection{Ablation Studies \& Other Evaluations}

Here, we discuss several additional experiments we performed to evaluate the performance of individual components of our system in isolation.

\begin{table}[t!]
    \centering
    \begin{tabular}{lccccc}
         \toprule
         \textbf{Synth. \& Trans.} & \textbf{Lines$\uparrow$} & \textbf{H$\downarrow$} & \textbf{M$\downarrow$} & \textbf{C$\downarrow$} & \textbf{U$\downarrow$} \\
         \midrule
         Combined (No stage 1) & 36.2 & \textbf{1.16} & 24.28 & 49.71 & 21.97 \\
         Ours (separated) & \textbf{49.4} & 9.73 & \textbf{16.22} & \textbf{31.63} & \textbf{14.6} \\
         \bottomrule
    \end{tabular}
    \caption{
    Evaluating how separating synthesis and translation into different LLM queries affects the complexity of the generated scene programs (Lines) as well as the rates at which the types of errors described in Section~\ref{sec:progsynth} occur (H = Hallucination, M = Misuse, C = Contradiction, U = Unsatisfiability).
    Our full pipeline improves all metrics but one (Hallucination).
    }
    \label{tab:progsynth_ablation}
\end{table}

\paragraph{Scene program synthesis}

In Section~\ref{sec:progsynth}, we discussed the benefits of splitting the LLM-based program synthesizer into stages which first generate a detailed natural language description of the scene and then translate that description into code.
Here, we empirically demonstrate those benefits.
For the prompts from the perceptual study in the previous section, 
we generate scenes using our full program synthesizer and a variant without the natural language description stage, i.e. in this variant, the LLM must synthesize the scene and translate it to code at the same time.
To assess the complexity of the generated scene programs, we measure the average number of lines per scene.
We also measure the frequency at which the four types of errors described in Section~\ref{sec:error_correction} occur.
Since scenes have a different number of objects \& relations (and thus a different number of chances to make errors), rather than report the average error rate per scene, we instead report the number of errors per 1000 objects.

Table~\ref{tab:progsynth_ablation} shows the results of this experiment.
Using our full pipeline results in more complex scene programs and leads to a reduction in the rates of all error types except hallucinations.
The higher rate of hallucinations in our pipeline is not surprising: since the first stage generates a free-form natural language description of the scene, the latter stages may ``invent'' new relation functions that correspond to parts of that description.
By contrast, such errors are less likely to happen when the LLM is instructed to directly generate a program with a fixed vocabulary of functions.
Nonetheless, the other benefits offered by separating synthesis and translation make this trade-off worth it.

\begin{figure}[t!]
    \centering
    \includegraphics[width=\linewidth]{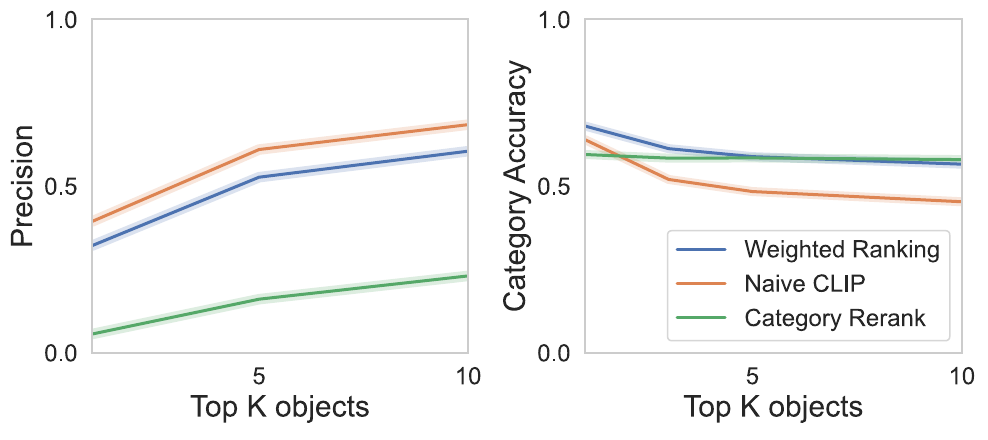}
    \caption{
    Plotting the retrieval precision (left) and category accuracy (right) of different ranking schemes for our open-universe object retrieval module (x axis is the number of top $k$ objects considered).
    Our weighted re-ranking approach preserves high category accuracy while incurring only a small hit to precision.
    }
    \label{fig:retrieval_eval}
\end{figure}

\paragraph{Object retrieval}
We first demonstrate the value of of our category-aware re-ranking scheme for object retrieval.
We compare this scheme to two alternatives: no re-ranking (naive retrieval), and re-ranking purely based on category (i.e. using only the second term in Equation~\ref{eq:weighted_rerank}).
We run each of these methods on text descriptions from the Cap3D dataset~\cite{cap3d}, which contains paired (3D mesh, text description) data with meshes sourced from Objaverse.
We compute the retrieval Precision@$K$ (how often is the ground-truth mesh associated with the text description contained in the top $K$ retrieved objects) as well as a similar metric of category accuracy (what percentage of the top $K$ retrieved objects are of the correct category).
Fig.~\ref{fig:retrieval_eval} shows plots for both of these metrics over increasing values of $K$.
Naive retrieval performs well in terms of precision, but suffers from a steep drop-off in category accuracy.
By contrast, re-ranking purely based on category leads to consistent category accuracy across $K$ at the cost of a significant hit to precision.
Our weighted re-ranking scheme achieves the best of both worlds: making sure that most of the top $K$ objects are of the correct category while retaining enough information about the overall description of the object to suffer only a minor hit to retrieval precision.

\begin{table}[t!]
    \centering
    \small
    \setlength{\tabcolsep}{3pt}
    \begin{tabular}{llcc}
        \toprule
        \textbf{Filter} & \textbf{Category} & \textbf{True Positive Rate $\uparrow$} & \textbf{False Positive Rate $\downarrow$}
        \\
        \midrule
        \multirow{5}{*}{\textit{Category}}
        & Bookcase & 0.77 & 0.27 \\
        & Rug & 0.88 & 0.22 \\
        & Painting & 0.97 & 0.1 \\
        & Table & 0.97 & 0.28 \\
        \cmidrule(lr){2-4}
        & Average & 0.90 & 0.21 \\
        \midrule
        \multirow{6}{*}{\textit{Multi-object}}
        & Desk & 0.75 & 0.15 \\
        & TV Stand & 0.95 & 0.17 \\
        & Side Table & 0.97 & 0.11 \\
        & Table & 0.92 & 0.2 \\
        & Couch & 0.89 & 0.31 \\
        \cmidrule(lr){2-4}
        & Average & 0.86 & 0.2 \\
        \bottomrule
    \end{tabular}
    \caption{
    Performance of our two object retrieval filters when used on a benchmark set of of objects with manually-labeled ground truth labels. Our filtration technique discards most unsuitable retrievals while retaining a large fraction of suitable retrievals.
    }
    \label{tab:retrieval_filters}
\end{table}

We also evaluate the performance of our two retrieval filters (category and multi-object).
For each filter, we chose a handful of common object categories and built benchmark datasets for each by producing a set of two object text descriptions and running them through the first part of our retrieval pipeline (initial retrieval and re-ranking).
For each set of retrieval results, we traverse the top $k$ objects and create a set containing 10 meshes which should pass the filter and 5 which should not (manually labeled by a human observer).
This results in a dataset with 802 annotated meshes. 
We then run our filters on all of these sets of meshes three times (to account for non-determinism in the LLM) and report their average true positive and false positive rates in Table~\ref{tab:retrieval_filters}.
Overall, both the category and multi-object filters achieve around a 90\% true positive rate and 20\% false positive rate.
Given the size of our 3D object dataset, this true positive rate is more than sufficient to ensure that enough valid candidate meshes can be retrieved in almost all cases.
This false positive rate means that roughly one in five meshes which passes a filter should actually have been rejected---not perfect, but a notable improvement over not filtering at all.

\begin{table}[t!]
    \centering
    \small
    \setlength{\tabcolsep}{1pt}
    \begin{tabular}{lcccccc}
         \toprule
         \textbf{Method} & \textbf{Chair} & \textbf{Couch} & \textbf{Desk} & \textbf{Wardrobe} & \textbf{Painting} & \textbf{Average} \\
         \midrule
         VLM (front) & 65.3 & 54.0 & 67.3 & 87.3 & 84.3 & 71.6 \\
         VLM (front,back) & 69.3 & 97.0 & 57.4 & 87.3 & 88.2 & 79.8 \\
         VLM (front,back,side) & \textbf{96.0} & 97.0 & 57.4 & 86.3 & 88.2 & 85.0 \\
         LLM & 75.0 & \textbf{97.4} & 92.4 & 94.1 & \textbf{94.8} & 90.7 \\
         Ours & 87.8 & 97.0 & \textbf{93.7} & \textbf{95.8} & \textbf{94.8} & \textbf{93.8} \\
         \bottomrule
    \end{tabular}
    \caption{
    Classification accuracies for different orientation prediction approaches when evaluated on a benchmark set of objects with manually-labeled ground truth orientations.
    Our approach combines a VLM with a multimodal LLM to get the best of both worlds, achieving the highest overall accuracy.
    }
    \label{tab:orientation}
\end{table}

\paragraph{Object orientation}
Finally, we evaluate the performance of our front-facing direction classifier.
Similarly to the previous experiment, we build a benchmark dataset (with 450 objects in total) containing 50-100 objects for each of several common categories, each of which has a hand-labeled ground-truth front facing direction.
We then evaluate how well our method for orientation prediction perform on this dataset, compared to the following alternatives we tried:
\begin{itemize}
    \item \textbf{Ours:} our full orientation prediction method as described in Section~\ref{sec:orientation}.
    \item \textbf{VLM (front):} choosing whichever orientation produces an image whose VLM embedding has the highest cosine similarity to that of the text ``the front of a $C$,'' where $C$ is the object category.
    \item \textbf{VLM (front,back):} like the previous method, but where we also render the reverse face of the object and add the similarity to ``the back of a $C$'' to the objective we minimize.
    \item \textbf{VLM (front,back,side):} like the previous method, but where we also render one of the side faces of the object and add the similarity to ``the side of a $C$'' to the objective we minimize.
    \item \textbf{LLM:} providing renders of all four faces of the object to an LLM and asking it to choose which image best represents the front of object.
\end{itemize}
Table~\ref{tab:orientation} shows the results of this experiment.
The method which only uses a multimodal LLM performs better than the VLM-based methods in general, but it does suffer from large performance drops on certain types of objects (e.g. chairs).
By using a VLM to filter the set of views the LLM must consider down to two (a task likely more prevalent in its training data than four-way image comparison), our method improves over the LLM-only baseline on nearly all categories.

\subsection{Timing}

On a MacBook Pro with an Apple M1 Max processor and 32GB RAM, the median time to generate an object layout is about four minutes.
Almost all of this time is spent querying the LLM (the layout optimizer stage takes under 10 seconds, typically).
Converting the layout into a full 3D scene is more computationally expensive, as our object retrieval and orientation modules can invoke multiple LLM calls for each object in the scene; for complex, densely-populated scenes, this cost adds up.
The median time to retrieve an object is slightly under a minute (51 seconds); to orient the object, it is 16 seconds.
For a set of scenes we generated with an average of 17 objects per scene, this led to a median total scene generation time of about 25 minutes.
While slower than prior systems for closed-universe scene synthesis (which often take only seconds), this is still faster than existing text-to-3D systems which optimize a VLM-based loss---these systems can take hours to produce a single scene. 
Our approach could also be accelerated by caching information computed about retrieved objects (e.g. their front-facing orientations, whether they belong to a certain category) to avoid re-computing those quantities if objects are encountered again.

%% file: 09-conclusion.tex
\section{Discussion \& Future Work}
\label{sec:conclusion}

We presented a system for open-universe scene generation: taking a text prompt as input, our system generates room-scale indoor scenes of any requested type composed of whatever relevant objects are needed for that room.
Our system leverages LLMs to generate scenes by tasking them with generating declarative object-relation programs; these programs are then converted to constraint problems which are solved with gradient-based optimization to produce object layouts.
Finally, our system uses multimodal LLMs and vision-language models to retrieve appropriate meshes for each object from a massive, unannotated dataset, as well as to estimate the front-facing orientation of these retrieved meshes.

Open-universe scene generation is a complex, challenging task, and our system for solving it is not perfect.
In the remainder of the paper, we discuss limitations and opportunities for improvement.

\subsection{Viability for Interior Design}
To get a sense for whether the outputs produced by our system could currently be used in real-world interior design scenarios, we conducted a small qualitative study.
We recruited six individuals (``clients'') seeking complimentary interior design services through online ads, gathered their design needs through 30-minute interviews, and later presented to them designs created by both our system and by professional designers.
Both the clients and the designers provided feedback on the generated scenes.
More detail about these interviews and about how the scenes were generated can be found in the supplemental material.

The clients and the designers appreciated the system's ability to produce appropriate groupings of objects (e.g. dining tables and chairs), correctly place rugs under furniture, and ensure adequate space for door openings.
They also appreciated the color and material coordination in its generated furniture objects.
However, they found that the system could produce overly cluttered scenes in which it seemed to lack an understanding of certain professional interior design principles such as maintaining circulation (by e.g. not clustering furniture in corners).
Such design principles can be expressed computationally~\cite{ifurniture_design}; our system could be improved by adding design principles as operations to our scene modeling DSL and allowing the program synthesis LLMs to decide which principles should be applied to which (parts of) scenes.

\subsection{Other Limitations \& Future Work}

Our system currently only supports four-walled rooms.
There are many ways this limitation could be removed: non-rectangular rooms could be subdivided into rectangular regions, or arbitrary arrangements of walls could be specified parametrically in the system's input prompt (though reasoning about the resultant wall geometry may prove difficult for an LLM; multimodal LLMs which can correlate parametric wall objects with images of wall geometry may help).
In addition, objects in the current system are restricted to one of four cardinal orientations.
This discrete set could be expanded.
Orientations could also be represented as continuous values in the layout optimizer (e.g. allowing small angular corrections to maintain \texttt{FACING} constraints); this would necessitate a revision to our current multi-step layout optimization scheme.

In very rare cases, objects in large databases such as Objaverse are modeled with their up axis not aligned with one of the world coordinate axes; this violates the assumptions of our orientation prediction module and can result in ``tilted'' objects being inserted into the scene.
It may be possible to detect and correct (or filter out) such objects using geometric heuristics~\cite{UprightOrientation} or carefully-designed queries to multimodal language models.

We have introduced some mechanisms for fixing errors produced by our LLM program synthesizer, but they are not foolproof.
In the future, we are interested in exploring LLM self-repair~\cite{LLMSelfCorrection} instead of / in addition to our existing heuristics: collecting detected errors and tasking the LLM with correcting its own prior output to eliminate them.

The open-ended capabilities of VLMs and LLMs could support myriad approaches for open-universe scene generation.
In this paper, we have explored one small region of this design space; future work is needed to map out its entirety.
We hope that our work serves as both a springboard and a strong baseline for a new line of research on open-universe scene generation.

%% file: 10-appendix.tex
\appendix

\section{Scene Description Language}
\label{sec:apndx_language}

As described in Section~\ref{sec:language}, the domain-specific functions we add to Python are either (1) object constructors, (2) relation functions or (3) parameter setting functions. The basic object constructor is
\begin{spverbatim}
Object(description: str, width: float, depth: float, height: float, facing: Object | int | None = None).

\end{spverbatim}
An object can face either one of the 4 cardinal directions (\texttt{EAST=0}, \texttt{NORTH=1}, \texttt{WEST=2}, \texttt{SOUTH=3}), another object, or not face anything. A programmer (a human or an LLM) may want not to specify a facing direction, if this direction is not important (for example, for a tablet lying on a sofa). In this case the object will appear in a scene in a random orientation.

In addition to the default \texttt{Object} constructor, we have special constructors for doors and windows:
\begin{spverbatim}
Door(description: str, width: float, height: float, wall: int)

Window(description: str, width: float, height: float, wall: int, height_above_ground: float, above: Cuboid = None)

\end{spverbatim}
Doors and windows are treated as regular objects in our system. The programmer can place additional relations on doors and windows. However, doors and windows initialize with additional constraints that ensure that (1) doors are always adjacent to walls, and windows are always mounted on walls, and (2) there is enough empty space in front of a door, so a door can be opened, windows also require empty space in front of them.

For some types of objects such as paintings, books or statues, it makes sense to retrieve different 3D meshes even for objects that have the same description. For this purpose, we have a list-of-objects constructor \texttt{unique\_objects(amount: int, description: str, width: float, depth: float, height: float)}. For consistency, we also have an \texttt{objects(amount: int, description: str, width: float, depth: float, height: float)} constructor, although it can be replaced with a list comprehension.

DSL relation functions are:
\begin{enumerate}
\item \texttt{on(top: Object, bottom: Object)} means that the first object stands on top of the bottom object. If the top object is smaller, it should not extend beyond the bottom object.

\item \texttt{next\_to\_wall(a: Object, wall: int, distance: float = 0.0)} means that object \texttt{a} stands next to one of the 4 walls. If the optional \texttt{distance} argument is zero, the object is touching the wall. Otherwise, it stands no more than \texttt{distance} meters from the wall. The optional distance parameter allows to fit a chair between a table and a wall when a table is standing next to a wall. Using a pair of \texttt{next\_to\_wall} functions, the programmer can express that some object stands in a corner.

\item \texttt{mounted\_on\_wall(a: Object, wall: int, height: float, above: Object = None)} means that that object \texttt{a} is mounted on a wall \texttt{height} meters above the ground. This relation is useful for paintings, mirrors, wall clocks or whiteboards. When the optional \texttt{above} argument is used, object \texttt{a} is mounted above some other object.

\item \texttt{mounted\_on\_ceiling(a: Object, above: Object = None)} means that object \texttt{a} is mounted on a ceiling, optionally above another object. This relation is useful for describing fans, projectors or chandeliers.

\item \texttt{adjacent(a: Object, b: Object, arg1: int | float | None = None, arg2: int | float | None = None, arg3: float = 0.0)} relation can be used with 0, 1 or 2 direction arguments, and an optional distance argument:
\begin{enumerate}
\item \texttt{adjacent(chair, desk)} means that the chair's bounding box touches the desk's bounding box.
\item \texttt{adjacent(chair, desk, NORTH)} is the most common variant. It means that the chair is adjacent to the desk from the NORTH.
\item \texttt{adjacent(chair, desk, NORTH, WEST)} means that chair is adjacent to the desk from the north side but is aligned with the west side of the desk. This version of the adjacency relation is useful for describing a nightstand that is adjacent to a bed from the side, but is aligned with the head of a bed.
\end{enumerate}
If an optional distance argument is used, the touching requirement is replaced with the distance requirement. For example, \texttt{adjacent(chair, desk, NORTH, 0.2)} means that the chair is located to the NORTH of the desk, no more than 20cm away from the desk.

\item \texttt{aligned(cuboids: list[Object], axis: bool)} means that the centers of a list of objects should be aligned either vertically or horizontally. The second argument can either be \texttt{WESTEAST=False} or \texttt{NORTHSOUTH=True}. This relation is useful for describing careful arrangements of furniture.

\item \texttt{facing(a: Object, direction: Object | int)} means that the forward vector of object \texttt{a} should face either one of the cardinal directions or another object.

\item \texttt{surround(chairs: list[Object], table: Object)} is a syntax sugar relation that is implemented with \texttt{adjacent} and \texttt{facing} relations. It adds surrounding objects (for example, chairs) one-by-one, and picks the sides of the central object (for a example, a table) that have the most free space available. This relation respects other adjacencies and walls.
\end{enumerate}

Finally, the DSL has parameter setting functions for setting the size of the scene, the floor texture and the wall texture. These functions are called only once per scene:
\begin{verbatim}
set_size(westeast: float, northsouth: float, height: float)
set_floor_texture(texture: str)
set_wall_texture(texture: str)
\end{verbatim}

\section{Layout Optimizer Details}
\label{sec:apndx_optimizer}

Here we provide more implementation details for our system's layout optimizer module.

\subsection{Constraint Losses}
Let $a^{size}$, $a^{center}$, $a^{min}$ and $a^{max}$ denote the size, the center, the lower-left-bottom corner and the upper-right-top corner of cuboid $a$, and let $s$ be the scene cuboid. Let $\relu(x)=\max(x,0)$ and let $\relu$ be defined for vectors component-wise. Let $\proj(v)$ be the projection of vector $v$ on the horizontal $XZ$-plane. Let $d(a,b)$ be Euclidean distance between cuboids $a$ and $b$. We define constraint losses in the following way:

\begin{equation*}
\mathcal{L}_{\text{WITHINBOUNDS}}(a) = \|\relu(s^{min}-a^{min})\|^2 + \|\relu(a^{max}-s^{max})\|^2,
\end{equation*}

\begin{equation*}\begin{split}
\mathcal{L}_{\text{ON}}(a,b) = (b^{max}_y-a^{min}_y)^2 + \|\relu(\proj(b^{min}-a^{min}))\|^2 +\\+ \|\relu(\proj(a^{max}-b^{max}))\|^2 - \frac 12\|\relu(\proj(a^{size}-b^{size}))\|^2,
\end{split}\end{equation*}

\begin{equation*}
\mathcal{L}_{\text{HEIGHT}}(a,height) = (a^{min}_y-height)^2,
\end{equation*}

\begin{equation*}
\mathcal{L}_{\text{NEXTTOWALL}}(a,\text{EAST}, dist) = \relu^2(s^{max}_x-a^{max}_x-dist),
\end{equation*}

\begin{equation*}
\mathcal{L}_{\text{CEILING}}(a) = (a^{max}_y-s^{max}_y)^2,
\end{equation*}

\begin{equation*}
\mathcal{L}_{\text{ALIGNED}}(a,b,\text{WESTEAST}) = (a^{center}_z-b^{center}_z)^2,
\end{equation*}

\begin{equation*}
\mathcal{L}_{\text{ABOVE}}(a,b,\text{EAST}) = \relu^2(B^{min}_z-A^{min}_z) + \relu^2(A^{max}_z-B^{max}_z),
\end{equation*}

where $A,B = a,b$ if $a^{size}_z<b^{size}_z$ and $A,B = b,a$ otherwise.

\begin{equation*}
\mathcal{L}_{\text{ADJACENT0}}(a,b,dist) = (d(a,b)-dist)^2,
\end{equation*}

\begin{equation*}\begin{split}
\mathcal{L}_{\text{ADJACENT1}}(a,b,\text{EAST},dist) = \mathcal{L}_{\text{A}}(a,b,\text{EAST},dist) +\\+ \relu^2(b^{min}_z-a^{min}_z),
\end{split}\end{equation*}

\begin{equation*}\begin{split}
\mathcal{L}_{\text{ADJACENT2}}(a,b,\text{EAST},\text{NORTH},dist) = \mathcal{L}_{\text{A}}(a,b,\text{EAST},dist) +\\+ (b^{min}_z-a^{min}_z)^2,
\end{split}\end{equation*}

where

\begin{equation*}\begin{split}
\mathcal{L}_{\text{A}}(a,b,\text{EAST},dist) = \relu^2(a^{min}_x-b^{max}_x-dist) +\\+ \relu^2(b^{max}_x-a^{min}_x) + \relu^2(a^{max}_z-b^{max}_z) -\\- \frac 12\relu^2(a^{size}_z-b^{size}_z).
\end{split}\end{equation*}

Generalization from \texttt{EAST} and \texttt{NORTH} to other cardinal directions is straightforward.

\subsection{Repel Force Implementation}

To implement repel forces, we first define a binary connectivity relation on the set of objects and walls in the scene: \texttt{NEXTTOWALL(a, wall)} connects object \texttt{a} and wall \texttt{wall}, and \texttt{ON(a,b)} or \texttt{ADJACENT(a, b)} connect objects \texttt{a} and \texttt{b}. This relation splits the set of objects and walls into connected components. For every pair of objects that belong to different connected components (and for pairs of objects and walls from different connected components) we add a repelling vector to the optimization gradient with magnitude proportional to $\max(1-d/d_\text{max}, 0)$, where $d$ is the Euclidean distance between the two objects (or between an object and a wall), and $d_\text{max}$ is the maximum range of repels, set to be the $1/4$ the minimum linear size of the scene. We add a small random noise to repel forces to escape local minima.